\documentclass{article}
\usepackage{iclr2027_conference,times}
\usepackage[T1]{fontenc}
\usepackage[utf8]{inputenc}
\usepackage{amsmath,amssymb}
\usepackage{graphicx}
\usepackage{hyperref}
\makeatletter
\newcommand{\InJepaCitationLink}[1]{%
  \mbox{\hyper@natlinkstart{\@citeb\@extra@b@citeb}#1\hyper@natlinkend}}
\def\NAT@hyper@#1{%
  \begingroup
    \let\InJepaCitationDate\NAT@date
    \let\InJepaCitationSuffix\NAT@exlab
    \def\NAT@date{\InJepaCitationLink{\InJepaCitationDate}}%
    \def\NAT@exlab{\InJepaCitationLink{\InJepaCitationSuffix}}%
    \let\hyper@natlinkbreak\@firstoftwo
    #1%
  \endgroup}
\makeatother
\usepackage{xurl}
\usepackage{float}
\usepackage{needspace}
\newsavebox{\jwjsectiondisplaybox}

\usepackage{capt-of}
\usepackage{array,booktabs}
\newcolumntype{N}[1]{>{\centering\arraybackslash}m{#1}}
\newcommand{\jwjrow}{\strut}
\newcommand{\jwjcompactrow}{\strut}
\newcommand{\jwjcell}[2]{#1\,#2}

\newcommand{\jwjtablebody}{%
  \normalsize
  \renewcommand{\arraystretch}{1}%
  \renewcommand{\jwjrow}{}%
  \renewcommand{\jwjcompactrow}{}%
}

\title{InJePA: Intention-First Latent Planning\\with Coupled JEPA World Models\\for Visual Navigation}
\author{%
Siming Zhou$^{1}$, Bowen Ma$^{1}$, Ruiran Du$^{1}$, Pengcheng Zheng$^{1}$, Sheng Zheng$^{1}$ \\
\textbf{Dongshen Han$^{1}$, Zhijie Li$^{2}$, Wei Dong$^{2}$, Fan Mo$^{1}$, Chaoning Zhang$^{1}$}\thanks{\raggedright Corresponding author: Chaoning Zhang (\href{mailto:chaoningzhang1990@gmail.com}{\texttt{chaoningzhang1990@gmail.com}}).\newline Contact information: Siming Zhou: \href{mailto:2024080901001@std.uestc.edu.cn}{\texttt{2024080901001@std.uestc.edu.cn}}.\newline Pengcheng Zheng: \href{mailto:zpc777@std.uestc.edu.cn}{\texttt{zpc777@std.uestc.edu.cn}}.} \\[3pt]
\textnormal{$^{1}$University of Electronic Science and Technology of China} \\
\textnormal{$^{2}$Xi'an University of Architecture and Technology}
}
\iclrfinalcopy
\hypersetup{pdftitle={InJePA: Intention-First Latent Planning with Coupled JEPA World Models for Visual Navigation},pdfauthor={Siming Zhou, Bowen Ma, Ruiran Du, Pengcheng Zheng, Sheng Zheng, Dongshen Han, Zhijie Li, Wei Dong, Fan Mo, Chaoning Zhang},pdfsubject={},
  pdfkeywords={Image Goal Navigation, World Models, Joint Embedding Predictive Architecture, Latent State Planning, Embodied AI}}

\begin{document}
\maketitle
% Remove the conference-publication header in this preprint.
\fancyhead{}
\renewcommand{\headrulewidth}{0pt}

\begin{abstract}
Visual navigation requires goal directed planning informed by predictions of environment dynamics.
Representative navigation world model methods propose complete action candidates before predicting consequences for selection or search updates.
For each candidate, predictions primarily evaluate proposed actions rather than guide its stepwise construction.
To this end, we propose InJepa, an intention first planning paradigm centered on latent states.
A goal conditioned JEPA proposes latent intentions, and an action conditioned JEPA continually predicts future latent states.
Both operate in a shared latent space to align their predictive representations.
Within each candidate, predicted consequences feed into the next intention proposal and action inference, forming a closed loop of intention, action, and consequence.
This extends the world model's role beyond evaluation to leading candidate construction through latent intentions.
We evaluate InJepa through closed loop image goal navigation in Habitat MP3D.
The results show effective multistep navigation through intention guided action generation and consequence feedback.
Experiments across encoders further support applicability across visual representations.
These findings support latent intentions as the central planning object and show that world models can directly guide plan construction through coordinated reasoning about intentions and consequences.
Code is available at \url{https://github.com/zhuxun132/InJepa}.
\end{abstract}

\section{Introduction}
Visual navigation is fundamental to autonomous robotics and embodied intelligence.
Agents must use visual observations to reach goals through continued interaction with the environment~\citep{chaplot2020neural,zeng2023moving,ko2024learning}.
In image goal navigation, reaching the goal from the current observation requires a sequence of state transitions.
Each action affects both the distance to the goal and the state for subsequent decisions.
Navigation must therefore connect visual goals with sequential decisions, accounting for both goal proximity and action consequences~\citep{ma2022vip}.
Reasoning about future state changes provides an important basis for coherent multistep planning and goal directed navigation~\citep{zhang2026world,zhang2026learning}.

Existing learned navigation methods typically generate actions through goal conditioned policies and use world model predictions to support planning.
These policies directly output action sequences from observation histories and goal images~\citep{shah2023vint,sridhar2024nomad}.
World models predict future observations, latent states, or compatibility between consequences and goals~\citep{hafner2019dream,hafner2020mastering,cheng2025scaling} to support candidate comparison~\citep{chua2018deep,hafner2019learning,wang2026predicting,hansen2024td,lu2025makes}.
In representative candidate search methods, a policy or optimizer first proposes complete action sequences.
The world model then predicts their consequences to select candidates or update the search distribution~\citep{zhang2026rae,chahe2026policy,bar2025navigation}.
Predictive modeling~\citep{zheng2026joint} has expanded from image generation to latent space~\citep{assran2023self,bardes2024revisiting,assran2025v,mur2026v,kuhn2026levjepa,maes2026leworldmodel}.
These methods evaluate candidate consequences before execution using different representations and computational forms to compare predicted outcomes with the goal.

\begin{figure}[!t]
\centering
% Crop only the old embedded headings; retain the uploaded artwork unchanged.
\includegraphics[trim=0 8bp 0 35bp,clip,width=0.90\linewidth]{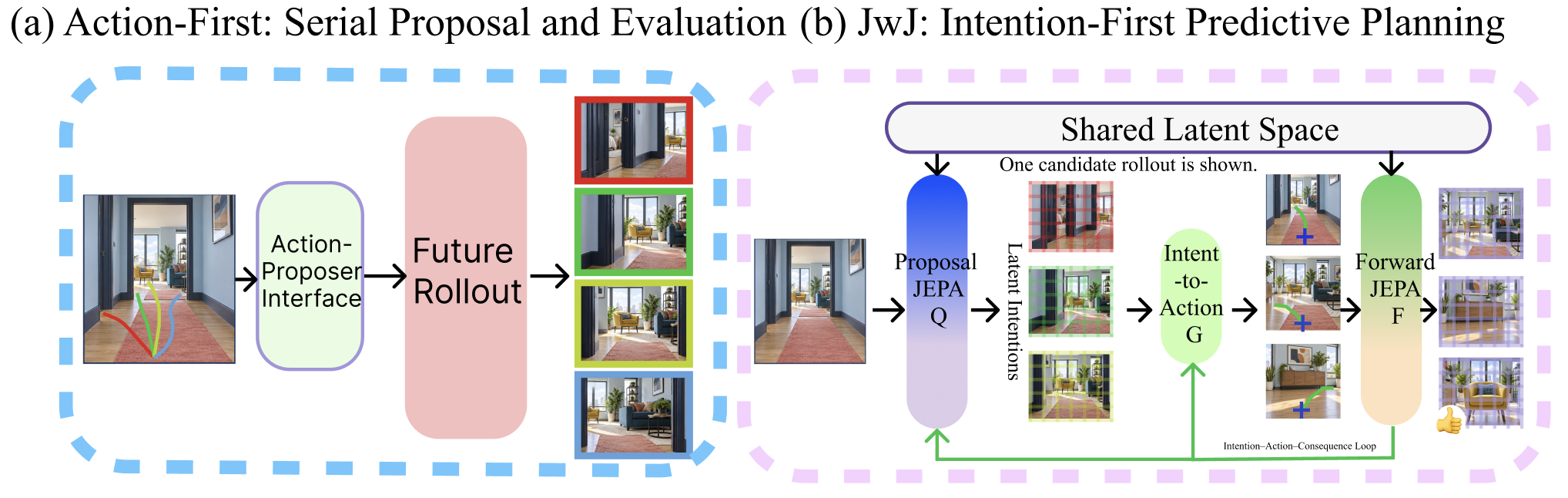}\par
{\small\makebox[0.45\linewidth][c]{(a) Action first paradigm}%
\makebox[0.45\linewidth][c]{(b) Our intention first architecture}\par}
\caption{From action first to intention first planning. (a) A policy or optimizer proposes complete action sequences for world model evaluation. (b) InJepa's world model constructs candidates through latent intention proposals guided by consequence feedback. Images illustrate latent states.}
\label{fig:motivation}
\end{figure}

However, this action first construction order limits the world model's role in forming each candidate.
Desired state changes toward the goal can only be determined indirectly through action candidates.
Although consequence evaluation can guide subsequent search, the current candidate's actions are fixed before prediction.
Predictions primarily evaluate proposed behaviors and do not participate in the stepwise generation of subsequent actions within that candidate.
We therefore reconsider the starting point of planning: \emph{can world models lead the stepwise construction of navigation plans by proposing desired changes and jointly reasoning about intentions and consequences?}

To address this issue, we propose InJepa, an intention first planning architecture.
We define an intention as a candidate desired latent state conditioned on the goal and couple intention proposals with consequence prediction in a shared latent space.
At each step, Proposal JEPA proposes an intention from the current state and goal.
The action model translates the difference between the intention and the current state into an action~\citep{sun2026intact}.
Forward JEPA then predicts the successor state under that action.
The predicted state becomes the common basis for subsequent intention generation and action inference within the same candidate.
As shown in Figure~\ref{fig:motivation}, the world model directly leads stepwise plan construction while retaining candidate evaluation.
Closed loop experiments in unseen MP3D scenes and evaluations with three visual encoders support the architecture's navigation effectiveness and applicability across representations.
Controlled mechanism comparisons further examine the roles of intention proposals and consequence feedback.

To conclude, the main contributions of this paper are as follows:
\begin{itemize}
    \item To the best of our knowledge, we propose the first intention first planning paradigm for world models in visual navigation.
    We shift the primary planning object from action sequences to goal conditioned latent intentions.
    World models thus lead stepwise plan construction through the generation and realization of desired state changes.

    \item We introduce InJepa, a dual JEPA architecture for joint reasoning over intentions and consequences in a shared latent space.
    Within each candidate, predicted successor states guide both subsequent intention generation and action inference, so plans are progressively constructed in response to the predicted consequences of earlier actions.

    \item We validate effective closed loop navigation in unseen MP3D scenes and applicability across three visual encoders.
    Through controlled mechanism comparisons and ablation studies, we further examine how intention proposals, consequence feedback, and training supervision contribute to effective navigation.
\end{itemize}
% End the introduction page without stretching the contribution-list spacing.
\newpage

\section{Related Work}

\subsection{Navigation World Models}
Navigation world models mainly use two forms of prediction.
One generates future images or videos and evaluates candidates by comparing predicted observations with the goal.
NWM predicts videos conditioned on actions~\citep{bar2025navigation}.
Subsequent work improves generation consistency and rollout efficiency~\citep{yan2026mwm,shen2026efficient,liu2026relaxing}.
It also explores joint or alternating generation of visual sequences and actions~\citep{chen2026pondering,dong12026towards}.
The other predicts and evaluates consequences in latent space.
RAE-NWM, PiJEPA, and ReL-NWM use feature predictions to guide action search~\citep{zhang2026rae,chahe2026policy,zhang2025efficient}.
LWM learns compatibility between consequences and goals for candidate selection and policy learning~\citep{wang2026predicting}.
Related work jointly models latent prediction and action generation~\citep{yang2026wam}, or uses compact representations to reduce rollout cost~\citep{kim2026planning,jiang2022efficient,zheng2026llava}.
Across both forms, action candidate methods plan directly over action sequences, determining desired changes indirectly through consequence evaluation or action optimization.

\subsection{Planning in State Space and Action Realization}
Guidance by anticipated effects and coordination between forward and inverse models provide cognitive and control foundations for planning in state space~\citep{kunde2002impact,wolpert1998multiple}.
One class decomposes tasks using latent landmarks or visual subgoals for execution by lower level policies~\citep{shah2021rapid,zhang2021world,moghaddam2021learning,nair2019hierarchical,bell2024language}.
FF-JEPA predicts latent subgoals and uses CEM to search for actions to realize them~\citep{masip2026ff}.
HWM generates subgoals through high level macro actions, then optimizes low level actions~\citep{zhang2026hierarchical}.
Another class learns mappings from plans to actions~\citep{lynch2020learning,thomas2023plex,ajay2022conditional,li2024efficient,gao2025flip}.
LDP decodes latent state trajectories into actions through inverse dynamics~\citep{xie2025latent}.
NavDreamer converts video plans into navigation waypoints~\citep{huang2026navdreamer}.
INTACT infers actions from the difference between the goal and current states.
It uses predicted consequences to update the state reference for the next action inference step~\citep{sun2026intact}.
These approaches establish state planning and predictive feedback~\citep{du2024video}, but separate subgoal execution, trajectory decoding, or high level proposal search from the coupled process studied here.
Our focus is the dependency within each candidate: a desired latent state determines an action, whose predicted consequence conditions both the next local intention and its action realization.

\begin{figure}[!t]
\centering
\includegraphics[trim=0 10bp 0 14bp,clip,width=0.60\linewidth]{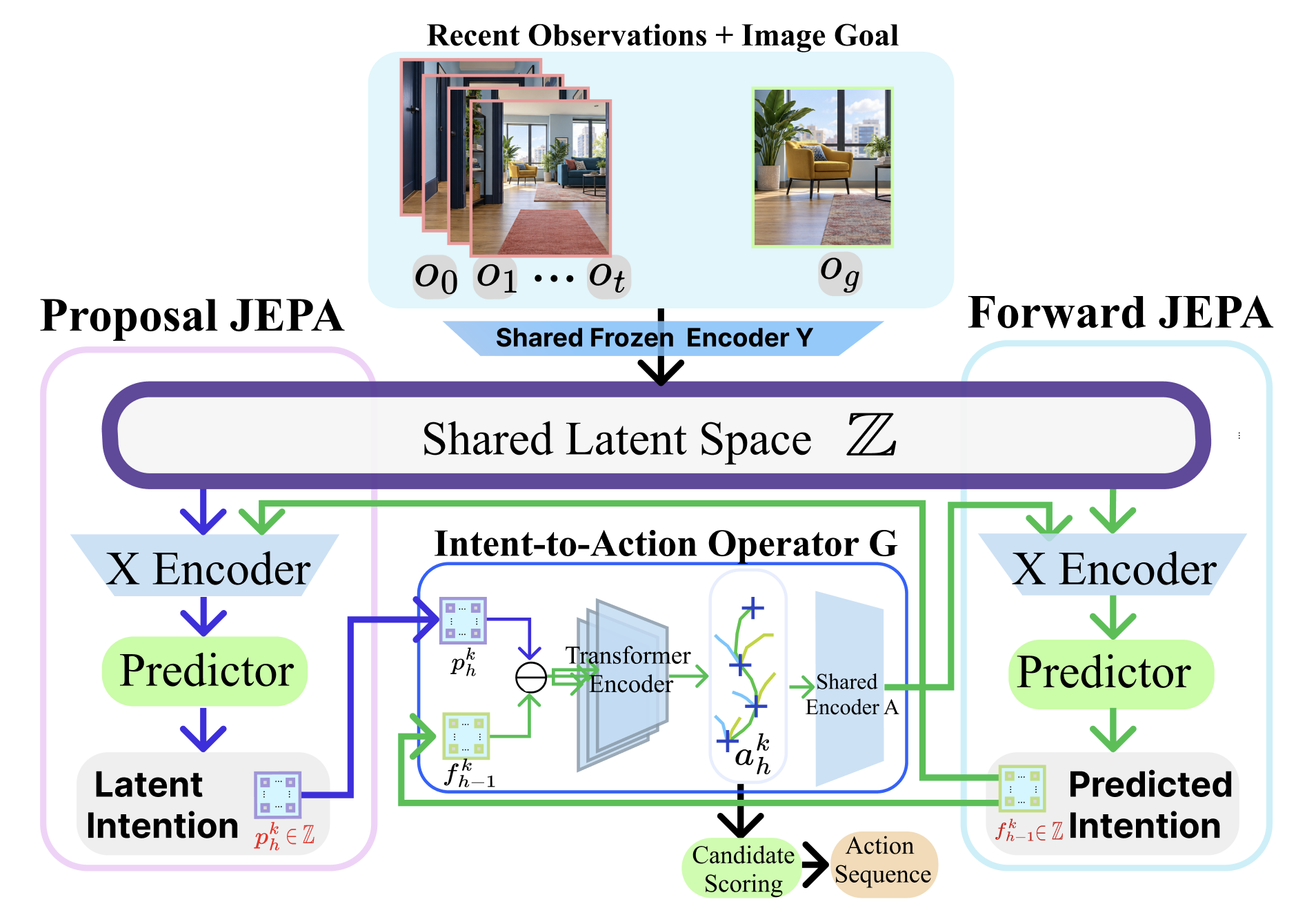}
\caption{Intention first planning architecture of InJepa. In a shared latent space, Proposal JEPA proposes intentions and the action model $G$ translates them into actions. Forward JEPA feeds predicted consequences into subsequent intention generation and action inference.}
\label{fig:architecture}
\end{figure}

\section{Method}

InJepa organizes planning around goal directed latent intentions, infers actions to realize desired changes, and feeds predicted consequences into subsequent intention generation and action inference.
Appendix~\ref{sec:implementation-details} details network and training settings, context masks, and deployment rules.

\subsection{Problem Formulation}

At environment step $t$, an image goal navigation agent selects an action $a_t\in\mathcal A$ from its recent observation and action history $\mathcal H_t$ and goal image $o_g$ to reach the goal location. The action set $\mathcal A$ contains forward, left, right, and stop. The first three form the movement action set $\mathcal A_{\mathrm{move}}$. A frozen V-JEPA 2.1 encoder $Y$ maps the current observation $o_t$ and goal into shared \linebreak[4] spatial feature grids~\citep{mur2026v}:
\[
z_t=Y(o_t),\qquad z_g=Y(o_g),\qquad
z_t,z_g\in\mathcal Z=\mathbb R^{S\times D},
\tag{1}
\]
where $z_t$ and $z_g$ are current and goal grids in $\mathcal Z$, with $S$ visual tokens and feature dimension $D$. The context $c_t$ consists of recent latent observations and their associated actions. The planner rolls out $K$ candidates from the current state, each for $H$ steps. Indices $k\in\{1,\ldots,K\}$ and $h\in\{1,\ldots,H\}$ denote candidates and rollout steps, distinct from the environment step $t$.

\subsection{Planning with Latent Intentions}

As shown in Figure~\ref{fig:architecture}, InJepa distinguishes the desired state $p_h^k$ from the predicted action consequence $f_h^k$, both in $\mathcal Z$, with $f_0^k=z_t$. Proposal JEPA $Q$ proposes a state target from the context and goal before the corresponding action is specified. $G$ translates the difference between the target and current state into an action, and Forward JEPA $F$ predicts its consequence. Both JEPAs predict target encodings in a shared representation space~\citep{assran2023self}.
The parameters of $Q$ (including memory and prior), $G$, and $F$ are $\phi$, $\theta$, and $\psi$, respectively.
An intention specifies the state sought through action. Consequence prediction estimates the resulting state under the selected action. The predicted state then guides both subsequent intention generation and action inference within the same candidate. Intention first planning thus specifies the order of plan construction at each step, without fixing an entire intention trajectory in advance.

\subsection{Goal Conditioned Intention Generation}

The memory encoder $M_\phi$ of $Q$ organizes recent visual grids, historical action encodings, and the goal grid into a condition memory $m_t=M_\phi(c_t,z_g)$. It uses only historical actions, without access to future actions under planning. The goal thus directly conditions intention proposals.
Under shared conditions, we sample each candidate variable $\xi^k$ from the conditional diagonal Gaussian prior~$P_\phi$~via~reparameterization~\citep{sohn2015learning,kingma2013auto}:
\[
P_\phi(\xi\mid c_t,z_g)
=\mathcal N\!\left(\mu_\phi(m_t),
\operatorname{diag}(\sigma_\phi^2(m_t))\right),
\;
\xi^k=\mu_\phi(m_t)+\sigma_\phi(m_t)\odot\epsilon^k,
\; \epsilon^k\sim\mathcal N(0,I),
\tag{2}
\]
where $\mu_\phi$ and $\sigma_\phi$ denote the conditional mean and standard deviation, respectively. Here $\mathcal N$ denotes a Gaussian distribution, $\epsilon^k$ is standard Gaussian noise, $I$ is the identity matrix, $\operatorname{diag}$ forms a diagonal matrix, and $\odot$ denotes elementwise multiplication. Different candidates use different samples, while $\xi^k$ remains fixed throughout each candidate rollout. The candidate variable controls intention generation. The resulting intention states serve as planning objects for action inference.

At each rollout step $h$, a shared recurrent unit generates the next intention from the current state, condition memory, and candidate variable:
\[
p_h^k=Q_\phi(f_{h-1}^k,m_t,\xi^k),
\qquad h=1,\ldots,H.
\tag{3}
\]
The first input is the real state $z_t$. Subsequent inputs are states $f_{h-1}^k$ predicted by $F$. The observation and goal conditions remain fixed within each planning round.

\subsection{Action Realization and Lookahead Selection}

To realize each intention, $G_\theta$ outputs an action distribution $\pi_\theta^G$ over $\mathcal A$ from the current predicted state, intention difference, and previous action. It selects the rollout action $a_h^k$ as
\[
a_h^k=\underset{a\in\mathcal A_{\mathrm{move}}}{\arg\max}
\ \pi_\theta^G\!\left(
 a\mid f_{h-1}^k,p_h^k-f_{h-1}^k,a_{h-1}^k
\right).
\tag{4}
\]
We set $a_0^k=a_{t-1}$ from real history. The same intention can map to different actions from different states.
$F$ predicts the successor from a rolling context of observed and predicted state grids $\mathcal C_{h-1}^k$ and aligned departure actions $\mathcal U_h^k$, with $a_h^k$ as the newest action~\citep{assran2025v}:
\[
f_h^k=F_\psi(\mathcal C_{h-1}^k,\mathcal U_h^k).
\tag{5}
\]
The predicted state $f_h^k$ becomes the next state input to both $Q$ and $G$.

Lookahead selection balances goal proximity and intention realization. For feature grids $x,y\in\mathcal Z$, define $d_1(x,y)=\|x-y\|_1/(SD)$ using the entrywise L1 norm, and let $\bar d_1$ denote this mean distance with a numerical floor. The relative intention realization error is
\[
r_h^k=\frac{d_1(p_h^k,f_h^k)}{\bar d_1(p_h^k,f_{h-1}^k)}.
\tag{6}
\label{eq:relative-realization-error}
\]
For negligible requested changes, we floor $r_h^k$ at 1. Numerical safeguards are provided in Appendix~\ref{sec:scoring-details}. For each complete $H$ step branch, the score $J_k$ combines normalized terminal goal distance and maximum relative error over its rollout:
\[
J_k=\frac{d_1(f_H^k,z_g)}{\bar d_1(z_t,z_g)}
+\lambda_{\mathrm c}\max_{1\leq h\leq H}r_h^k.
\tag{7}
\label{eq:candidate-score}
\]
The weight $\lambda_{\mathrm c}$ balances goal proximity and relative realization error. Candidates have equal weights and contain only movement actions. We select $k^\star=\arg\min_k J_k$, execute $a_1^{k^\star}$, and replan from a new real observation~\citep{assran2025v}. Arrival is defined in Section~\ref{sec:experimental-setup}.
Section~\ref{sec:mechanism-reasoning} examines consequence feedback and selection with fixed weights. Algorithm~1 in Appendix~\ref{sec:planning-algorithm} summarizes candidate rollout, selection, and execution.

\subsection{Learning Objectives and Training}

Offline trajectories provide no explicit intention labels. The recorded next observation $o_{t+1}$ provides $z_{t+1}=Y(o_{t+1})$ to supervise $Q$ and $F$, while recorded actions supervise $G$. We also train $G$ to map both real state differences and $Q$ proposals to actions. $\mathbb E$ averages over valid training samples and posterior draws, and $\operatorname{sg}$ denotes stop gradient.

\subsubsection{Goal Conditioned Intentions}

We learn a goal conditioned distribution of single step state proposals from recorded next states using conditional variational learning~\citep{sohn2015learning}. During planning, sampled proposals serve as candidate targets for action realization. A conditional diagonal Gaussian posterior $R_\eta(\xi\mid c_t,z_g,z_{t+1})$ with trainable parameters $\eta$ provides a sample $\widetilde\xi\sim R_\eta$ for the training intention $\widetilde p=Q_\phi(z_t,m_t,\widetilde\xi)$. The proposal loss $\mathcal L_Q$ combines reconstruction with KL regularization to align the posterior with the prior conditioned only on the context and goal:
\[
\mathcal L_Q=\mathbb E\!\left[
\ell_{\mathrm{rec}}(\widetilde p,z_{t+1})
+\ell_{\mathrm{KL}}(R_\eta,P_\phi)
\right].
\tag{8}
\label{eq:proposal-loss}
\]
Here $\ell_{\mathrm{rec}}$ is the mean Laplace negative log likelihood, and $\ell_{\mathrm{KL}}$ is the weighted, normalized Kullback--Leibler regularization term. Detailed loss definitions are provided in Appendix~\ref{sec:training-configuration}. Inference uses only the prior and recursively rolls out multistep candidates with predicted consequences.

\subsubsection{Action Consequence Prediction}

To model action consequences, we train $F$ to predict the next state grid $\widehat z_{t+1}$ from the real history and executed action. We minimize the mean squared error $\mathcal L_F$ between this prediction and the frozen next state grid~\citep{zhou2024dino}:
\[
\mathcal L_F=\mathbb E\left[
\frac{1}{SD}\|\widehat z_{t+1}-\operatorname{sg}(z_{t+1})\|_F^2
\right].
\tag{9}
\]
Here $\|\cdot\|_F$ denotes the Frobenius norm.
\subsubsection{Intention to Action Mapping}

To support intention realization, we train $G$ to map state differences to actions using the dual branch supervision of INTACT~\citep{sun2026intact}. The two branches use differences between adjacent real states and between the trajectory endpoint grid $z_T$ and current state, respectively. Here $T$ indexes the terminal state. The corresponding action losses are
\[
\begin{aligned}
\mathcal L_{G,\mathrm{local}}
&=\mathbb E[-\log\pi_\theta^G(a_t\mid z_t,z_{t+1}-z_t,a_{t-1})],\\
\mathcal L_{G,\mathrm{goal}}
&=\mathbb E[-\log\pi_\theta^G(a_t\mid z_t,z_T-z_t,a_{t-1})].
\end{aligned}
\tag{10}
\]
The goal branch also includes terminal states with zero state differences and STOP labels. To adapt $G$ to intention inputs at inference, the loss $\mathcal L_{QG}$ supervises action inference from the posterior proposal $\widetilde p$ with recorded actions:
\[
\mathcal L_{QG}=\mathbb E[-\log\pi_\theta^G(a_t\mid z_t,
\operatorname{sg}(\widetilde p)-z_t,a_{t-1})].
\tag{11}
\label{eq:qg-loss}
\]
This loss stops gradients at $\widetilde p$ and trains only $G$ to interpret intention proposals. $Q$ remains supervised by the latent prediction objective in Equation~\eqref{eq:proposal-loss}. The total objective is
\[
\mathcal L=\mathcal L_Q+\mathcal L_F
+\lambda_{\mathrm{local}}\mathcal L_{G,\mathrm{local}}
+\lambda_{\mathrm{goal}}\mathcal L_{G,\mathrm{goal}}
+\lambda_{QG}\mathcal L_{QG}.
\tag{12}
\]
The weights $\lambda_{\mathrm{local}}$, $\lambda_{\mathrm{goal}}$, and $\lambda_{QG}$ scale local, goal, and intention conditioned action supervision, respectively. The visual encoder remains frozen. $\mathcal L_Q$ supervises $Q$ and its prior and posterior, $\mathcal L_F$ supervises $F$, and the three action losses supervise $G$. The corresponding branches update the action embedding shared by $G$ and $F$. $Q$ reads historical action embeddings with stopped gradients. At inference, consequence feedback updates rollout inputs without changing model parameters.

\section{Experiments}

\subsection{Experimental Setup}
\label{sec:experimental-setup}

\subsubsection{Implementation Details}
\label{sec:experimental-implementation}
We jointly train $Q$, $G$, and $F$ on spatial features from a frozen visual encoder.
Unless stated otherwise, InJepa denotes InJepa-VJEPA with $K=8$, $H=4$, and $\lambda_{\mathrm c}=1$ in Equation~\eqref{eq:candidate-score}.
To assess applicability across representations, we replace only the visual encoder and retrain to obtain InJepa-DINOv3~\citep{simeoni2025dinov3} and InJepa-CroCo.
Each encoder has approximately 86M parameters.
We train each V-JEPA/DINOv3 variant on four RTX 4090 GPUs and CroCo on four RTX 3090 GPUs.
Architecture and training details are provided in Appendix~\ref{sec:implementation-details}.

\subsubsection{Benchmarks}
\label{sec:experimental-benchmarks}
Training and offline diagnostics use the public StreamVLN R2R/RxR RGB and action trajectories~\citep{wei2025streamvln}, collected in Habitat~\citep{savva2019habitat} MP3D~\citep{chang2017matterport3d}.
The disjoint training and validation sets cover 50 and 11 buildings, respectively.
We fix 4,096 validation origin and goal pairs for offline diagnostics across training stages and ablated models.
We evaluate image goal navigation in Habitat MP3D on Clean150: 150 fixed tasks in 11 scenes unseen during training and validation.
The navigation distance scale and success radius follow RAE-NWM~\citep{zhang2026rae}.
The mean initial shortest geodesic distance is 7.93\,m. Each task allows at most 200 environment control steps, with arrival defined as reaching within 1\,m of the goal.
Each method uses its discrete action or waypoint interface within the same control step budget.

\begin{table}[!t]
    \centering
    \caption{Closed loop navigation performance. SR: Reach SR (\%). SPL: ReachSPL.}
    \label{tab:navigation-results}
    \jwjtablebody
    \setlength{\tabcolsep}{0.5pt}
    \begin{tabular*}{\linewidth}{@{\extracolsep{\fill}}l|*{3}{N{28pt}N{28pt}|}N{52pt}N{67pt}@{}}
        \toprule
        & \multicolumn{2}{c|}{Seed 0}
        & \multicolumn{2}{c|}{Seed 1}
        & \multicolumn{2}{c|}{Seed 2}
        & \multicolumn{2}{c@{}}{Mean $\pm$ SD} \\
        \cmidrule(lr){2-3}\cmidrule(lr){4-5}\cmidrule(lr){6-7}\cmidrule(l){8-9}
        Method
        & SR $\uparrow$ & SPL $\uparrow$
        & SR $\uparrow$ & SPL $\uparrow$
        & SR $\uparrow$ & SPL $\uparrow$
        & SR $\uparrow$ & SPL $\uparrow$ \\
        \midrule
        \jwjrow NoMaD
        & 29.33 & 0.2052 & 26.67 & 0.1733 & 28.67 & 0.1900
        & \jwjcell{28.22}{$\pm$\,1.39}
        & \jwjcell{0.1895}{$\pm$\,0.0159} \\
        \jwjrow RAE-NWM
        & 13.33 & 0.0522 & 15.33 & 0.0610 & 11.33 & 0.0470
        & \jwjcell{13.33}{$\pm$\,2.00}
        & \jwjcell{0.0534}{$\pm$\,0.0070} \\
        \midrule
        \jwjrow LWM-CroCo
        & \textbf{71.33} & \textbf{0.6866} & \textbf{69.33} & \textbf{0.6743} & \textbf{70.00} & \textbf{0.6731}
        & \jwjcell{\textbf{70.22}}{$\pm$\,\textbf{1.02}}
        & \jwjcell{\textbf{0.6780}}{$\pm$\,\textbf{0.0075}} \\
        \jwjrow LWM-VJEPA
        & 43.33 & 0.4018 & 42.00 & 0.3838 & 42.00 & 0.3539
        & \jwjcell{42.44}{$\pm$\,0.77}
        & \jwjcell{0.3798}{$\pm$\,0.0242} \\
        \midrule
        \jwjrow\textbf{InJepa-DINOv3 (ours)}
        & \textbf{50.00} & \textbf{0.4518} & \textbf{56.00} & \textbf{0.5143} & \textbf{53.33} & \textbf{0.4921}
        & \jwjcell{\textbf{53.11}}{$\pm$\,\textbf{3.01}}
        & \jwjcell{\textbf{0.4861}}{$\pm$\,\textbf{0.0317}} \\
        \jwjrow\textbf{InJepa-CroCo (ours)}
        & \textbf{62.00} & \textbf{0.5618} & \textbf{62.00} & \textbf{0.5814} & \textbf{61.33} & \textbf{0.5577}
        & \jwjcell{\textbf{61.78}}{$\pm$\,\textbf{0.38}}
        & \jwjcell{\textbf{0.5670}}{$\pm$\,\textbf{0.0127}} \\
        \jwjrow\textbf{InJepa-VJEPA (ours)}
        & \textbf{68.00} & \textbf{0.6357} & \textbf{62.67} & \textbf{0.5951} & \textbf{65.33} & \textbf{0.6132}
        & \jwjcell{\textbf{65.33}}{$\pm$\,\textbf{2.67}}
        & \jwjcell{\textbf{0.6147}}{$\pm$\,\textbf{0.0203}} \\
        \bottomrule
    \end{tabular*}
\end{table}

\begin{figure}[!t]
    \centering
    \centerline{\includegraphics[trim=0 14bp 0 6bp,clip,width=0.80\linewidth]{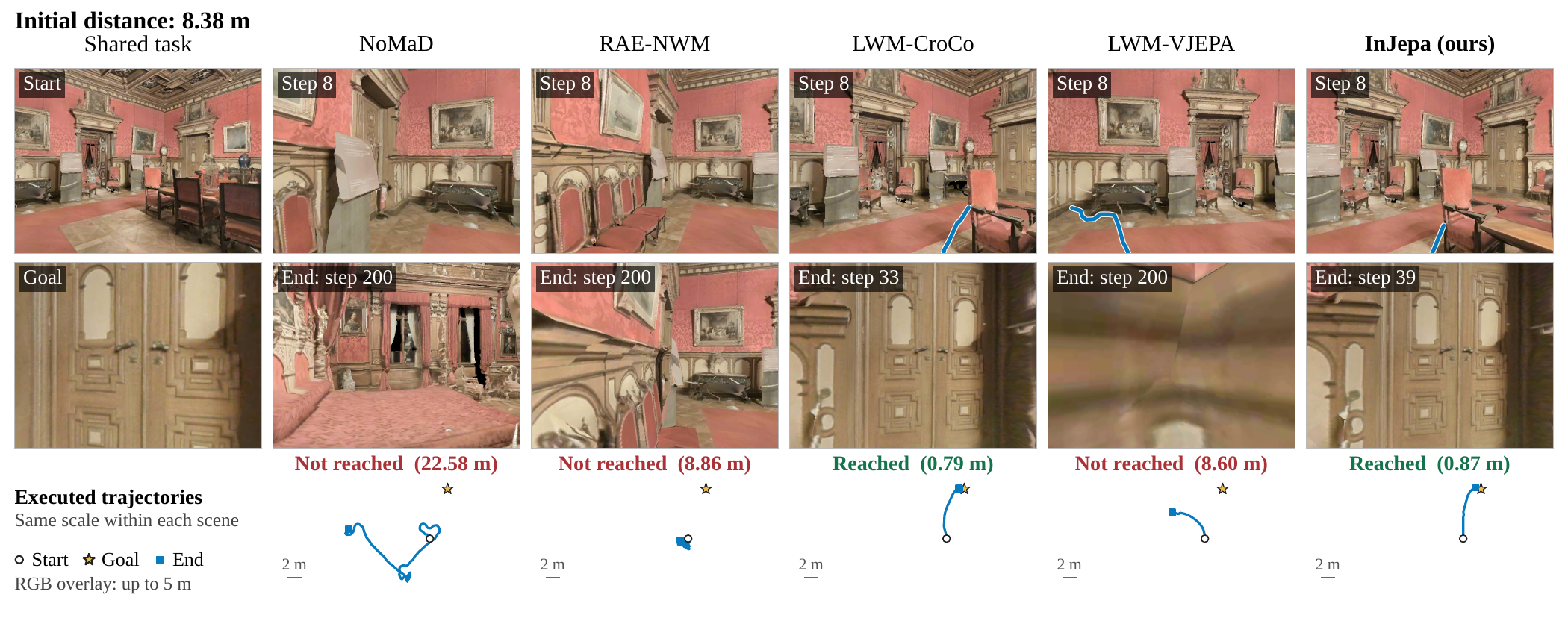}}
    \caption{Closed loop navigation in an unseen scene. All methods share the start and goal. Blue curves show executed trajectories. Distances are geodesic distances to the goal.}
    \label{fig:navigation-qualitative}
\end{figure}

\subsubsection{Baselines and Metrics}
\label{sec:baselines-and-metrics}
We compare direct policies conditioned on goals and navigation methods based on world models:
\begin{itemize}
    \item NoMaD~\citep{sridhar2024nomad} is a goal conditioned diffusion policy.
    Official weights provide a reference independent of our StreamVLN retraining pipeline.
    We disable map modules and pass the goal image directly to the policy.

    \item We reproduce RAE-NWM~\citep{zhang2026rae} on the same StreamVLN data, retaining its action conditioned latent dynamics, CEM search, and goal feature scoring.
    We only reduce Euler discretization from 50 to 10 points to lower its high planning cost. Appendix~\ref{sec:euler-discretization} reports the original setting's cost and completed evaluations.

    \item LWM~\citep{wang2026predicting} is, to our knowledge, the strongest current navigation world model.
    Using its official implementation and the same StreamVLN data, we reproduce LWM-CroCo with the original CroCo-v2~\citep{weinzaepfel2023croco} and LWM-VJEPA with V-JEPA~2.1~\citep{mur2026v}, changing only the visual encoder.
\end{itemize}

\paragraph{Metrics.}
\textbf{Reach SR ($\uparrow$)} is the fraction of tasks reaching the goal within the interaction budget. \textbf{ReachSPL ($\uparrow$)} additionally weights success by path efficiency.
Offline feature errors use RMS distance and aggregate valid samples at each step $h\in\{1,2,3,4\}$:
\begin{itemize}
    \item \textbf{Q Future Alignment ($\downarrow$).} The minimum feature error between candidate intentions and the recorded future measures coverage of the reference future.

    \item \textbf{Pairwise Diversity.} Mean RMS distance between distinct Q candidates measures diversity.

    \item \textbf{F Prediction Error ($\downarrow$).} Autoregressive feature error against recorded futures under recorded actions measures consequence prediction accuracy and error accumulation.

    \item \textbf{G Local / G Goal ($\uparrow$).} Top 1 accuracy against recorded actions assesses inference from the observed next state (G Local) or goal (G Goal), each relative to the current state.

    \item \textbf{Intent Realization Error ($\downarrow$).} Feature error between Q intentions and F consequences on the selected branch measures intention realization consistency within the model.
\end{itemize}
Repeat counts accompany results. Metric definitions and aggregation are provided in Appendix~\ref{sec:evaluation-metrics}.

\subsection{Main Results and Analysis}
\label{sec:closed-loop-navigation}

\subsubsection{Quantitative Comparison}
\label{sec:quantitative-comparison}
All three InJepa variants outperform NoMaD~\citep{sridhar2024nomad}, RAE-NWM~\citep{zhang2026rae}, and LWM-VJEPA~\citep{wang2026predicting} in SR and SPL (Table~\ref{tab:navigation-results}), supporting effective intention first navigation in unseen scenes.
LWM-CroCo performs best but declines noticeably with V-JEPA~2.1~\citep{mur2026v}, suggesting that its advantage is related to the fit between CroCo representations~\citep{weinzaepfel2023croco} and compatibility evaluation, potentially aided by explicit training for candidate scoring.
With the same V-JEPA representation, InJepa outperforms LWM-VJEPA and approaches LWM-CroCo, supporting more effective use of visual information through intention guided action generation and consequence feedback.
The encoder variants retain the primary V-JEPA hyperparameters without systematic optimization for each encoder. Their effective navigation and relatively limited performance differences support applicability across V-JEPA~2.1, DINOv3, and CroCo.

\subsubsection{Qualitative Visualization}
\label{sec:qualitative-visualization}
In Figure~\ref{fig:navigation-qualitative}, InJepa passes the furniture through local heading adjustments, with every forward action reducing goal distance. This is consistent with local intention guidance and consequence feedback during planning. LWM-CroCo maintains progress, whereas LWM-VJEPA loses its initial progress, consistent with the representation sensitivity in the quantitative comparison. NoMaD takes substantial detours after approaching the goal, indicating difficulty sustaining goal guidance through direct action generation in this case. RAE-NWM travels 34.52\,m but remains within 1.74\,m of the start. Repeated local motion consumes most of its budget, showing that latent prediction and CEM search do not sustain spatial progress in this scene.

\begin{table}[!t]
    \centering
    \caption{Planning mechanism ablations. Proposal only omits F. Post hoc F uses F for scoring. $F\to G$ and Full add feedback to G and Q+G, respectively. Action-first InJepa omits Q and uses G's goal branch with F feedback and scoring. Details are provided in Appendix~\ref{sec:deployment-mechanisms}.}
    \label{tab:mechanism-results}
    \jwjtablebody
    \setlength{\tabcolsep}{2pt}
    \begin{tabular*}{\linewidth}{@{\extracolsep{\fill}}l|*{3}{N{27.5pt}N{27.5pt}|}N{52pt}N{67pt}@{}}
        \toprule
        & \multicolumn{2}{c|}{Seed 0}
        & \multicolumn{2}{c|}{Seed 1}
        & \multicolumn{2}{c|}{Seed 2}
        & \multicolumn{2}{c@{}}{Mean $\pm$ SD} \\
        \cmidrule(lr){2-3}\cmidrule(lr){4-5}\cmidrule(lr){6-7}\cmidrule(l){8-9}
        Deployment
        & SR $\uparrow$ & SPL $\uparrow$
        & SR $\uparrow$ & SPL $\uparrow$
        & SR $\uparrow$ & SPL $\uparrow$
        & SR $\uparrow$ & SPL $\uparrow$ \\
        \midrule
        \jwjrow Proposal only
        & 32.67 & 0.2426 & 34.67 & 0.2775 & 28.00 & 0.2377
        & \jwjcell{31.78}{$\pm$\,3.42}
        & \jwjcell{0.2526}{$\pm$\,0.0217} \\
        \jwjrow Post hoc F
        & 58.00 & 0.5359 & 55.33 & 0.5096 & 54.67 & 0.4921
        & \jwjcell{56.00}{$\pm$\,1.76}
        & \jwjcell{0.5126}{$\pm$\,0.0220} \\
        \jwjrow $F\to G$
        & 62.67 & 0.5728 & 56.67 & 0.5151 & 59.33 & 0.5395
        & \jwjcell{59.56}{$\pm$\,3.01}
        & \jwjcell{0.5425}{$\pm$\,0.0289} \\
        \midrule
        \jwjrow $F{\to}Q{+}G$ \textbf{(full)}
        & \textbf{68.00} & \textbf{0.6357} & \textbf{62.67} & \textbf{0.5951} & \textbf{65.33} & \textbf{0.6132}
        & \jwjcell{\textbf{65.33}}{$\pm$\,\textbf{2.67}}
        & \jwjcell{\textbf{0.6147}}{$\pm$\,\textbf{0.0203}} \\
        \midrule
        \jwjrow Action-first InJepa
        & 48.67 & 0.4073 & 40.00 & 0.3302 & 40.00 & 0.3368
        & \jwjcell{42.89}{$\pm$\,5.00}
        & \jwjcell{0.3581}{$\pm$\,0.0428} \\
        \bottomrule
    \end{tabular*}
\end{table}

\begin{table}[!t]
    \centering
    \caption{Intention alignment and realization. Validation seed 0, four steps weighted equally. Q alignment uses the best candidate and realization uses the selected branch.}
    \label{tab:mechanism-offline}
    \jwjtablebody
    \setlength{\tabcolsep}{2pt}
    \begin{tabular*}{\linewidth}{@{\extracolsep{\fill}}l|ccccc|ccccc@{}}
        \toprule
        & \multicolumn{5}{c|}{Q Future Alignment $\downarrow$}
        & \multicolumn{5}{c@{}}{Intent Realization Error $\downarrow$} \\
        \cmidrule(lr){2-6}\cmidrule(l){7-11}
        Deployment
        & $h=1$ & $h=2$ & $h=3$ & $h=4$ & Mean
        & $h=1$ & $h=2$ & $h=3$ & $h=4$ & Mean \\
        \midrule
        Proposal only
        & \textbf{0.4057} & 0.5266 & 0.5881 & 0.6179 & 0.5346
        & 0.1837 & 0.3283 & 0.3896 & 0.4520 & 0.3384 \\
        Post hoc F
        & \textbf{0.4057} & 0.5266 & 0.5881 & 0.6179 & 0.5346
        & 0.1745 & 0.2893 & 0.3550 & 0.4171 & 0.3090 \\
        $F\to G$
        & \textbf{0.4057} & 0.5266 & 0.5881 & 0.6179 & 0.5346
        & 0.1735 & 0.2827 & 0.3467 & 0.4060 & 0.3022 \\
        \midrule
        $F{\to}Q{+}G$ \textbf{(full)}
        & \textbf{0.4057} & \textbf{0.5234} & \textbf{0.5796} & \textbf{0.6041} & \textbf{0.5282}
        & \textbf{0.1450} & \textbf{0.2047} & \textbf{0.2477} & \textbf{0.2916} & \textbf{0.2222} \\
        \bottomrule
    \end{tabular*}
\end{table}

\subsection{Ablation Studies}
\label{sec:ablation-studies}

\subsubsection{Planning Mechanisms}
\label{sec:mechanism-reasoning}

Tables~\ref{tab:mechanism-results} and~\ref{tab:mechanism-offline} compare planning mechanisms with fixed weights. Proposal only exceeds NoMaD~\citep{sridhar2024nomad} and RAE-NWM~\citep{zhang2026rae} in mean SR and SPL (Table~\ref{tab:navigation-results}), supporting effective goal directed action generation from Q's intentions. Post hoc F improves navigation and reduces realization error on the selected branch, indicating greater consistency between intentions and consequences within the model. Feedback to G provides smaller further gains, consistent with compensating for discrepancies between intended and predicted states. Full feedback achieves the highest SR and SPL and lowest mean alignment and realization errors. The first three variants leave Q unchanged, so their alignment errors are identical. Full feedback instead updates Q's subsequent proposals from predicted successor states, supporting a contribution from improving intention generation. Additional $K/H$ results are provided in Appendix~\ref{sec:additional-mechanism-results} (Tables~\ref{tab:mechanism-other-configurations}--\ref{tab:mechanism-k4-repeats}).

Action-first InJepa generates actions directly from the final goal and retains F for consequence feedback and candidate selection, yet its SR and SPL are lower even than Post hoc F (Table~\ref{tab:mechanism-results}). This further supports the effectiveness of Q's intention generation and its role in specifying desired state changes before action generation. Compared with direct use of the final goal, Q's candidate desired latent states provide G with local action targets suited to the task. G only needs to translate them into actions. Together with the gains from full feedback, these results support organizing candidate plans around latent intentions and using predicted consequences to continually update subsequent intentions and actions within each candidate.

\begin{figure}[!t]
    \centering
    \centerline{\includegraphics[trim=0 1.2bp 0 1.2bp,clip,width=\linewidth]{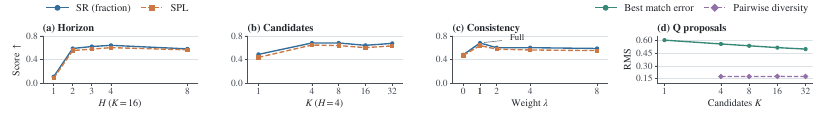}}
    \caption{Planning configuration. Navigation (a--c) and Q coverage and diversity (d), seed 0. Diagnostics use means over four steps. SR is a fraction.}
    \label{fig:planning-configuration}
\end{figure}

\begin{figure}[!t]
    \centering
    \centerline{\includegraphics[trim=0 1.2bp 0 1.2bp,clip,width=\linewidth]{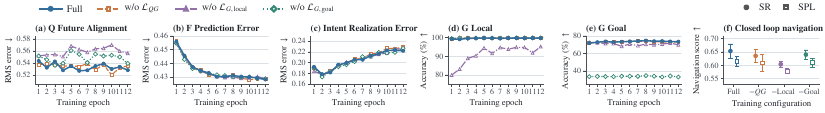}}
    \caption{Training supervision. Validation over epochs 1--12 (a--e) and epoch 12 navigation (f). Errors averaged over four steps. Navigation mean $\pm$ SD over three seeds, with SR as a fraction.}
    \label{fig:training-supervision}
\end{figure}

\subsubsection{Planning Configuration}
\label{sec:planning-configuration-ablation}
With $K=16$, navigation improves up to $H=4$ but declines at $H=8$, supporting multistep intention rollout with limited lookahead (Figure~\ref{fig:planning-configuration}(a)).
With $H=4$, multiple candidates outperform one, but larger $K$ does not consistently improve navigation (Figure~\ref{fig:planning-configuration}(b)).
Increasing $K$ improves reference coverage while mean pairwise distance remains largely stable, indicating broader sampling coverage rather than better individual intentions (Figure~\ref{fig:planning-configuration}(d)).
Moderate counts suffice: $K=8$ slightly outperforms $K=4$ on average across three seeds (Tables~\ref{tab:mechanism-results} and~\ref{tab:mechanism-k4-repeats}).
The consistency weight $\lambda_{\mathrm c}=1$ yields higher SR and SPL than no penalty or larger weights (Figure~\ref{fig:planning-configuration}(c)). Candidate evaluation therefore requires balancing goal proximity and intention realization, as excessive consistency penalties also limit navigation gains.

\subsubsection{Training Supervision}
\label{sec:training-supervision-ablation}
Each ablation sets one of $\lambda_{\mathrm{local}}$, $\lambda_{\mathrm{goal}}$, or $\lambda_{QG}$ to zero, retaining the other weights.
Removing local action supervision reduces G Local accuracy and causes the largest navigation decline, supporting accurate local state-to-action mappings for intention realization (Figure~\ref{fig:training-supervision}(d,f)).
Removing QG supervision leaves G Local accuracy largely unchanged but increases realization error and reduces mean navigation performance (Figure~\ref{fig:training-supervision}(c,d,f)). Action inference from observed transitions therefore cannot fully replace QG supervision for matching intentions to actions.
Removing goal action supervision reduces G Goal accuracy noticeably but changes closed loop performance only slightly, distinguishing direct goal conditioned action inference from intention guided navigation (Figure~\ref{fig:training-supervision}(e,f)).
F prediction errors remain similar across configurations (Figure~\ref{fig:training-supervision}(b)).
The full model attains the highest mean SR and SPL and lowest mean intention alignment and realization errors (Figure~\ref{fig:training-supervision}(a,c,f)), supporting complementary supervision for goal guidance, local action realization, and adaptation to intentions.

\Needspace{5\baselineskip}
\section{Conclusion and Future Work}
\label{sec:conclusion}

We present InJepa, an intention first paradigm for visual navigation.
In a shared latent space, it proposes goal directed intentions, translates them into actions, and feeds predicted consequences into subsequent intention generation and action inference.
Experiments validate effective navigation in unseen scenes and support applicability across representations and the benefits of consequence feedback.
Future work will study how to reduce the impact of accumulated prediction errors on longer horizon planning and assess the architecture's applicability in dynamic environments.

\subsection*{Reproducibility statement}

Appendix~\ref{sec:implementation-details} specifies architecture, training, hardware, and deployment. Appendix~\ref{sec:encoder-reproduction} details the encoder variants. Appendix~\ref{sec:baseline-deployment} documents baseline reproduction, data adaptations, controls, and compute. Appendices~\ref{sec:data-evaluation-details} and~\ref{sec:evaluation-metrics} specify data processing, task construction and poses, parameter selection, diagnostics, and metric aggregation.

\bibliography{references_v2}
\bibliographystyle{iclr2027_conference}
\clearpage
\appendix
% Continue after the twelve explicitly tagged equations in the main text.
\setcounter{equation}{12}
\section{Additional Mechanism Results}
\label{sec:additional-mechanism-results}

All results cover the complete set of 150 tasks.
The comparisons use the same model weights, IID sampling, one action per replanning cycle, and no stagnation protection.
SR denotes Reach SR (\%), and SPL denotes ReachSPL.
Tables~\ref{tab:mechanism-other-configurations} and~\ref{tab:mechanism-additional-runs} report seed 0 results.
Incomplete evaluations are excluded.

\begin{table}[!t]
    \centering
    \caption{Planning mechanisms under additional $K/H$ configurations (seed 0).}
    \label{tab:mechanism-other-configurations}
    \jwjtablebody
    \begin{tabular*}{\linewidth}{@{\extracolsep{\fill}}l|N{38pt}N{38pt}|N{38pt}N{38pt}@{}}
        \toprule
        & \multicolumn{2}{c|}{$K=4,H=4$}
        & \multicolumn{2}{c@{}}{$K=16,H=4$} \\
        \cmidrule(lr){2-3}\cmidrule(l){4-5}
        Deployment
        & SR $\uparrow$ & SPL $\uparrow$
        & SR $\uparrow$ & SPL $\uparrow$ \\
        \midrule
        \jwjcompactrow Proposal only & 39.33 & 0.3352 & 18.00 & 0.1498 \\
        \jwjcompactrow Post hoc F & 60.67 & 0.5427 & 57.33 & 0.5130 \\
        \jwjcompactrow $F\to G$ & 55.33 & 0.5050 & 58.67 & 0.5295 \\
        \midrule
        \jwjcompactrow $F\to Q+G$ (full) & \textbf{68.00} & \textbf{0.6453} & \textbf{64.00} & \textbf{0.6006} \\
        \bottomrule
    \end{tabular*}
\end{table}

Table~\ref{tab:mechanism-additional-runs} lists the remaining completed configurations.
These configurations do not yet have complete comparisons across all four mechanisms.
We report only the deployment variants with completed evaluations.

\begin{table}[!t]
    \centering
    \caption{Additional completed planning configurations. One evaluation per entry (seed 0).}
    \label{tab:mechanism-additional-runs}
    \jwjtablebody
    \begin{tabular*}{\linewidth}{@{\extracolsep{\fill}}N{36pt}N{36pt}|l|N{52pt}N{52pt}@{}}
        \toprule
        $K$ & $H$ & Deployment & SR $\uparrow$ & SPL $\uparrow$ \\
        \midrule
        \jwjcompactrow 16 & 1 & $F\to G$ & 10.67 & 0.0817 \\
        \jwjcompactrow 4 & 2 & $F\to Q+G$ & 61.33 & 0.5895 \\
        \jwjcompactrow 8 & 2 & $F\to Q+G$ & 58.67 & 0.5594 \\
        \jwjcompactrow 16 & 3 & $F\to G$ & 60.67 & 0.5712 \\
        \jwjcompactrow 16 & 3 & $F\to Q+G$ & 62.00 & 0.5762 \\
        \jwjcompactrow 1 & 4 & $F\to Q+G$ & 48.67 & 0.4354 \\
        \jwjcompactrow 32 & 4 & $F\to Q+G$ & 67.33 & 0.6309 \\
        \jwjcompactrow 16 & 8 & $F\to Q+G$ & 58.00 & 0.5639 \\
        \bottomrule
    \end{tabular*}
\end{table}

Full feedback at $K=4,H=4$ has been evaluated with three seeds, while the other mechanisms have only seed 0 results.
Table~\ref{tab:mechanism-other-configurations} therefore uses the common seed 0 for the complete mechanism comparison.
Table~\ref{tab:mechanism-k4-repeats} reports the additional repeats and their summary separately.

\begin{table}[!t]
    \centering
    \caption{Full feedback at $K=4,H=4$: three seeds, mean $\pm$ sample standard deviation.}
    \label{tab:mechanism-k4-repeats}
    \jwjtablebody
    \setlength{\tabcolsep}{2.5pt}
    \begin{tabular*}{\linewidth}{@{\extracolsep{\fill}}l|*{3}{N{27.5pt}N{27.5pt}|}N{52pt}N{67pt}@{}}
        \toprule
        & \multicolumn{2}{c|}{Seed 0}
        & \multicolumn{2}{c|}{Seed 1}
        & \multicolumn{2}{c|}{Seed 2}
        & \multicolumn{2}{c@{}}{Mean $\pm$ SD} \\
        \cmidrule(lr){2-3}\cmidrule(lr){4-5}\cmidrule(lr){6-7}\cmidrule(l){8-9}
        Deployment
        & SR $\uparrow$ & SPL $\uparrow$
        & SR $\uparrow$ & SPL $\uparrow$
        & SR $\uparrow$ & SPL $\uparrow$
        & SR $\uparrow$ & SPL $\uparrow$ \\
        \midrule
        \jwjrow $F\to Q+G$
        & 68.00 & 0.6453 & 62.00 & 0.5924 & 61.33 & 0.5916
        & \jwjcell{63.78}{$\pm$\,3.67}
        & \jwjcell{0.6098}{$\pm$\,0.0308} \\
        \bottomrule
    \end{tabular*}
\end{table}

\section{Evaluation Metrics}
\label{sec:evaluation-metrics}

\subsection{Closed Loop Navigation}

For task $i$, let $s_i$ indicate whether the agent reaches within the goal radius during the interaction budget, $\ell_i$ be the initial shortest geodesic distance, and $L_i$ be the executed path length.
For $N$ tasks, we report
\begin{equation}
\label{eq:reach-sr}
\mathrm{Reach\ SR}=\frac{1}{N}\sum_{i=1}^{N}s_i,
\end{equation}
\begin{equation}
\label{eq:reach-spl}
\mathrm{ReachSPL}=\frac{1}{N}\sum_{i=1}^{N}
s_i\frac{\ell_i}{\max(\ell_i,L_i)}.
\end{equation}
Failed tasks contribute zero to both metrics. A task already within the goal radius at initialization receives a ReachSPL score of one.
Arrival uses a geodesic distance below 1\,m.
Reach SR measures goal reaching, while ReachSPL additionally penalizes unnecessary travel.
Tables report Reach SR as a percentage, while the ablation plots use a fraction as indicated in their captions.

\subsection{Offline Diagnostics}

We evaluate feature errors on native spatial grids.
For grids $u,v\in\mathbb R^{S\times D}$, their RMS distance is
\begin{equation}
\label{eq:diagnostic-rms}
d_{\mathrm{RMS}}(u,v)=\sqrt{\frac{1}{SD}\sum_{s=1}^{S}\sum_{j=1}^{D}(u_{sj}-v_{sj})^2}.
\end{equation}
Prediction metrics are aggregated over valid samples separately for $h\in\{1,2,3,4\}$.
This diagnostic distance differs from the normalized $L_1$ consistency penalty used for candidate selection in Equation~\eqref{eq:candidate-score}.

\paragraph{Q Future Alignment.}
Given observation history and a goal image, Q samples from its prior and recursively generates candidate intentions.
At each step, we take the minimum distance between the candidate intentions and the corresponding recorded future features, then average over valid samples.
The best candidate can differ across prediction steps.
This measures candidate coverage of the reference future.
Multiple plausible futures can lead to the same goal, so we do not require all candidates to match one recorded trajectory.

\paragraph{Pairwise Diversity.}
At each prediction step, we average RMS distances over all unordered pairs of Q candidates:
\begin{equation}
\label{eq:pairwise-diversity}
D_h=\frac{2}{K(K-1)}\sum_{1\leq k<j\leq K}d_{\mathrm{RMS}}(p_h^k,p_h^j),\qquad K>1.
\end{equation}
We then average over valid samples.
The metric is undefined for $K=1$ and measures candidate diversity as $K$ varies.

\paragraph{F Prediction Error.}
Starting from real observation history, F predicts autoregressively under the recorded action sequence.
Each predicted state enters the rollout context. Real future observations serve only as evaluation references.
Distances to the corresponding recorded future features measure prediction accuracy and error accumulation.

\paragraph{G Local / G Goal.}
G Local conditions action inference on the difference between the observed next state and current state.
G Goal instead uses the difference between the goal state and current state.
Both report Top 1 accuracy against recorded actions, assessing local transition and goal conditioned action inference, respectively.
These diagnostics use observed states or goals as conditions, rather than Q proposals.

\paragraph{Intent Realization Error.}
We roll out and select candidates under the corresponding deployment mechanism.
At each step of the selected branch, we measure the RMS distance between the Q intention and F consequence, then average over valid samples.
For proposal only, F predicts the consequences of the selected actions solely for this diagnostic. It does not affect action generation or candidate selection.
Lower error indicates greater consistency within the model. It does not itself measure arrival in the environment.

\paragraph{Aggregation and repeated evaluations.}
The four-step summaries in the tables and figures weight the four per-step means equally.
The number of repeated evaluations accompanies each result.
When multiple evaluation seeds are available, summaries report the arithmetic mean and sample standard deviation across runs.

\Needspace{18\baselineskip}
\section{Planning Procedure}
\label{sec:planning-algorithm}

\noindent\textbf{Algorithm 1: One intention first planning decision.}
\begin{enumerate}
    \item Build $c_t,m_t,z_t,z_g$ from the real observation and action history and goal.
    \item Sample $K$ candidate variables $\xi^{1:K}$ with Equation~(2). Set $f_0^k=z_t$ and initialize each candidate context from the real history.
    \item For $h=1,\ldots,H$, roll out all candidates in parallel. Use Equation~(3) to propose an intention, Equation~(4) to infer an action, and Equation~(5) to predict its consequence. Then update the candidate context.
    \item Compute the score $J_k$ of each complete branch with Equation~\eqref{eq:candidate-score}, including the near-zero request safeguard. Select the lowest score, breaking ties by candidate index, and execute $a_1^{k^\star}$.
    \item Receive the next real observation, update the factual history, and discard all imagined states. Repeat within the interaction budget. Success requires geodesic distance to the goal below 1\,m.
\end{enumerate}

\section{Implementation Details}
\label{sec:implementation-details}

\subsection{Network Configuration}

\begin{table}[!t]
\centering
\caption{Network dimensions. All visual encoders are frozen, and spatial grids retain their native resolution.}
\label{tab:network-configuration}
\begingroup
\jwjtablebody
\begin{tabular}{@{}ll@{}}
\toprule
Component & Configuration \\
\midrule
Visual feature / hidden width & 768 / 768 \\
V-JEPA 2.1 / DINOv3 / CroCo grids & $24\times24$ / $16\times16$ / $14\times14$ \\
Q decoder / G encoder / F decoder blocks & 6 / 3 / 6 \\
Attention heads / FFN width / dropout & 16 / 2048 / 0.1 \\
Candidate latent dimension $L$ & 128 \\
Observation context length & 4 \\
Unique trainable parameters (Q, A, G, F) & 135,235,064 \\
Prior / posterior MLP widths & $768\to768\to256$ / $1536\to768\to256$ \\
Action embedding MLP widths & $4\to3072\to768$ \\
G input projection / classifier & $3072\to768$ / $768\to4$ \\
F output MLP widths & $768\to2048\to768$ \\
\bottomrule
\end{tabular}
\endgroup
\end{table}

\paragraph{Intention representation and decoder.}
The candidate variable $\xi^k\in\mathbb R^L$ has dimension $L=128$. Its Gaussian noise and prior and posterior statistics have the same dimension. $Q$ forms $m_t$ from four recent visual grids, their incoming action embeddings, and the goal grid. Tokens carry fixed two dimensional sinusoidal positions, sinusoidal relative ages, and learned record type embeddings. The memory contains $4(S+1)+S$ tokens, with invalid history masked. A learned query reads the condition memory through attention and an FFN to form the prior summary. A separate readout summarizes the observed next grid for the posterior. The prior head reads the condition summary, while the posterior head reads both summaries. Each head uses LayerNorm, a linear layer, GELU, and a linear output for mean and scale parameters.

At each rollout step, a $768\to768$ projection of the current state is added to spatial positions, a query type embedding, and a $128\to768$ projection of $\xi^k$. Six Transformer blocks apply spatial self attention, cross attention to $m_t$, and an FFN~\citep{vaswani2017attention}. Each block also receives a learned projection of $\xi^k$. LayerNorm and a linear projection produce the intention grid. The cell is shared across steps and has no rollout index embedding. The factual condition memory and each candidate latent remain fixed within one planning round.

\paragraph{Action inference.}
Following INTACT~\citep{sun2026intact}, $G$ concatenates the current state, intention difference, their elementwise product, and the broadcast previous action embedding. A linear projection, three spatial Transformer blocks, and a learned action token feed a four class classifier. The shared action encoder applies a $4\to4$ pointwise convolution followed by the MLP in Table~\ref{tab:network-configuration}, with SiLU activation. A zero action vector represents the beginning of a trajectory.

\paragraph{Consequence prediction and context validity.}
$F$ uses six Transformer decoder blocks over spatial state queries and a memory of recent state and action tokens. Its output MLP uses LayerNorm and GELU. Each update retains the four most recent states and their aligned departure actions. The newest departure action is $a_h^k$. Histories shorter than four states are right aligned and masked. Imagined states remain within the current candidate and are discarded after execution. Neither pooling nor additional whitening is applied to the feature grids.

\paragraph{Numerical parameterization of Q.}
Q uses bounded LayerNorm gain $2\tanh(\gamma/2)$ and variance offset $10^{-4}$. In each attention head of width $d_a=48$, queries and keys are divided by $\sqrt{\|x\|_2^2+d_a10^{-4}}$. Attention logits use a learned temperature bounded by $2\sqrt{d_a}$ and initialized to $0.3072\sqrt{d_a}$. Values are scaled as $v/\sqrt{1+\operatorname{mean}(v^2)/4}$. Decoder block $\ell$ uses residual scale $1/\sqrt{4\ell}$, while the summary readouts use $1/\sqrt{2}$.

For raw prior head outputs $u,v$, we set $\mu_P=5\tanh(u/5)$ and $\log\sigma_P=(\log2)\tanh(v/\log2)$. Raw posterior outputs $u',v'$ define $\delta=4\tanh(u'/4)$ and $\omega=(\log2)\tanh(v'/\log2)$, giving $\mu_R=\mu_P+\sigma_P\delta$ and $\log\sigma_R=\log\sigma_P+\omega$. The prior output layer starts at zero, yielding a standard normal prior. Posterior output weights have standard deviation $10^{-3}$ and zero bias. These controls are shared across encoder and loss variants.

\subsection{Training Configuration}
\label{sec:training-configuration}

The terms in Equation~\eqref{eq:proposal-loss} are
\begin{equation}
\label{eq:proposal-reconstruction-detail}
\ell_{\mathrm{rec}}(\widetilde p,z_{t+1})
=\frac{1}{SD}\sum_{s=1}^{S}\sum_{d=1}^{D}
\left(\frac{|\widetilde p_{sd}-z_{t+1,sd}|}{b}+\log(2b)\right),
\end{equation}
\begin{equation}
\label{eq:proposal-kl-detail}
\ell_{\mathrm{KL}}(R_\eta,P_\phi)
=\frac{\beta}{L}D_{\mathrm{KL}}\!\left(
R_\eta(\xi\mid c_t,z_g,z_{t+1})
\,\|\,P_\phi(\xi\mid c_t,z_g)\right).
\end{equation}
Here $s$ and $d$ index visual tokens and feature channels, $b=1/\sqrt{2}$ is the fixed Laplace scale, $\beta=0.05$ is the KL weight, and $D_{\mathrm{KL}}$ denotes Kullback--Leibler divergence. Reconstruction and KL are normalized by the number of grid elements $SD$ and candidate dimensions $L$, respectively.

\begin{table}[!t]
\centering
\caption{Training configuration. Loss ablations remove one action supervision term and retain the other weights.}
\label{tab:training-configuration}
\begingroup
\jwjtablebody
\begin{tabular}{@{}ll@{}}
\toprule
Setting & Value \\
\midrule
Q / F loss weights & 1 / 1 \\
$\lambda_{\mathrm{local}}$ / $\lambda_{\mathrm{goal}}$ / $\lambda_{QG}$ & 0.1 / 0.05 / 0.1 \\
KL weight $\beta$ / Laplace scale $b$ & 0.05 / $1/\sqrt{2}$ \\
Q supervised horizon / goal sampling horizon & 1 / 4 \\
Posterior samples per training occurrence & 1 \\
Optimizer & AdamW \\
Adam coefficients / numerical offset & $(0.9,0.999)$ / $10^{-8}$ \\
Weight decay & $10^{-3}$ \\
Peak learning rate: Q / shared A, G, F & $2\times10^{-4}$ / $3\times10^{-4}$ \\
Gradient norm clipping: Q / shared A, G, F & 1 / 1, separately \\
Global batch size / precision / training seed & 128 / BF16 mixed / 3072 \\
GPU processes / local batch / accumulation & 4 / 32 / 1 \\
Data workers / prefetch batches per worker & 8 / 2, per process \\
Learning rate schedule & Linear warmup, then cosine decay to zero \\
Schedule length / warmup fraction & 30 epochs / 1\% of planned updates \\
Reported navigation checkpoints & End of epoch 12 \\
\bottomrule
\end{tabular}
\endgroup
\end{table}

All trainable modules are initialized from scratch. Only the visual encoders use pretrained weights. Loss ablations also train from scratch, rather than continuing from the full model. We evaluate the checkpoint at the end of epoch 12 for each InJepa variant. The learning rate schedule retains its original 30 epoch duration without compression to 12 epochs. The primary run has 5,003 updates per epoch and 150,090 planned updates, including 1,500 warmup updates. Its epoch 12 checkpoint follows 60,036 updates. The three navigation seeds vary evaluation sampling, not training initialization. Each loss averages its own valid training occurrences before the weighted sum. QG gradients stop at the sampled posterior intention, as specified in Equation~\eqref{eq:qg-loss}.

The full model uses all five loss terms. The three loss ablations set only $\lambda_{QG}$, $\lambda_{\mathrm{local}}$, or $\lambda_{\mathrm{goal}}$ to zero, respectively. Other loss weights and the optimizer schedule remain unchanged. Encoder variants retain the same trainable architecture and principal optimization settings.

\paragraph{Hardware and software.}
The primary V-JEPA model and the DINOv3 variant were trained with four RTX 4090 GPUs. The training server provides 48\,GB of memory per GPU. The CroCo variant was trained with four RTX 3090 GPUs. All use global batch 128 across four processes, with no gradient accumulation. Activation checkpointing is enabled. The primary run retains activations in the final Q block and all G/F blocks, while the encoder variants use checkpointing throughout these blocks. The training runtime uses PyTorch 2.6 with CUDA 12.4. Habitat evaluation uses a separate Python 3.9 environment with PyTorch 2.5.1, CUDA 12.1, TorchVision 0.20.1, NumPy 1.26.4, and Habitat-Lab/Sim 0.2.4. Training and evaluation environments are distinct, and their CUDA builds should not be conflated. Evaluation was performed on RTX 3090 and RTX 4090 hardware.

\subsection{Candidate Scoring and Execution}
\label{sec:scoring-details}

Equation~\eqref{eq:candidate-score} selects complete branches of equal length $H$, not candidate prefixes. Its floored distance is $\bar d_1(x,y)=\max\{d_1(x,y),\epsilon\}$, with numerical floor $\epsilon=10^{-6}$. For clarity, with $u_h^k=d_1(p_h^k,f_{h-1}^k)$ and $v_h^k=d_1(p_h^k,f_h^k)$, the exact relative error is
\begin{equation}
\label{eq:relative-error-exact}
r_h^k=\begin{cases}
v_h^k/u_h^k,&u_h^k>\epsilon,\\
\max\{v_h^k/\epsilon,1\},&u_h^k\leq\epsilon.
\end{cases}
\end{equation}
The lower bound prevents a negligible requested change from receiving a spuriously small consistency cost. The goal term is evaluated only at $H$ and divided by the initial goal distance with the same numerical floor. The consistency term is the maximum over all $H$ steps, not their mean. The default $\lambda_{\mathrm c}=1$ applies to the primary model. Figure~\ref{fig:planning-configuration}(c) varies this same coefficient over $\{0,1,2,4,8\}$.

All $K$ sampled branches participate with equal weights. Ties are broken by candidate index. We execute one movement action, acquire a real observation, and initialize the next planning round from factual history. A task succeeds when geodesic distance to the goal falls below 1\,m. Simulator goal distance is used only by the evaluator, not as a planning input.

\subsection{Deployment Mechanisms}
\label{sec:deployment-mechanisms}

\begin{table}[!t]
\centering
\caption{Deployment definitions. All rows share model weights and initialization $p_0^k=f_0^k=z_t$.}
\label{tab:deployment-definitions}
\begingroup
\jwjtablebody
\begin{tabular}{@{}llll@{}}
\toprule
Mechanism & Q state input & G state and intention difference & Selection cost \\
\midrule
Proposal only & $p_{h-1}^k$ & $p_{h-1}^k,\ p_h^k-p_{h-1}^k$ & $d_1(p_H^k,z_g)$ \\
Post hoc F & $p_{h-1}^k$ & $p_{h-1}^k,\ p_h^k-p_{h-1}^k$ & $J_k$ \\
$F\to G$ & $p_{h-1}^k$ & $f_{h-1}^k,\ p_h^k-f_{h-1}^k$ & $J_k$ \\
$F\to Q+G$ & $f_{h-1}^k$ & $f_{h-1}^k,\ p_h^k-f_{h-1}^k$ & $J_k$ \\
\bottomrule
\end{tabular}
\endgroup
\end{table}

In each row of Table~\ref{tab:deployment-definitions}, Q also reads the fixed memory $m_t$ and candidate variable $\xi^k$, and G reads the previous generated action (the real previous action at $h=1$). Proposal only recursively generates intentions and translates their transitions into actions without invoking F. Post hoc F first fixes those actions, then rolls them out through F solely for scoring. With $F\to G$, the intention trajectory remains unchanged, but G uses the predicted current state at each step. Full feedback also supplies this state to the next Q call. Whenever F is used, it updates its own state and action history autoregressively.

The last three mechanisms use exactly the same $J_k$ and consistency weight. Thus their differences isolate the feedback path rather than a change in the scoring rule. Proposal only instead selects by its final intention. Its offline realization diagnostic uses a separate F rollout of the selected actions, without changing the policy. All mechanisms execute only the first selected action and replan.

Action-first InJepa reuses the Full epoch 12 checkpoint with $K=8$ and $H=4$, but does not call Q. At step $h$, G receives $f_{h-1}^k$, the goal difference $z_g-f_{h-1}^k$, and the previous action. Actions are sampled independently per branch from G's categorical distribution at temperature one. F predicts the successor and updates the history for the next G call. All sampled branches are retained, including duplicates, and ranked by $d_1(f_H^k,z_g)/\bar d_1(z_t,z_g)$ with the same numerical floor and index tie breaking. The intention consistency term is inapplicable because Q is absent. This control retains consequence feedback during action generation. It does not construct a complete action sequence before invoking F. It executes one action per planning round. Q alignment and intention realization diagnostics are not defined for this control. The comparison evaluates the intention based deployment as a whole. It does not isolate Q from the associated sampling and consistency scoring changes.

\subsection{Encoder Variants and Artifact Identity}
\label{sec:encoder-reproduction}

The DINOv3 and CroCo variants replace the frozen visual encoder and retrain Q, A, G, and F from scratch. Except for the spatial grids in Table~\ref{tab:encoder-interfaces}, the Q/A/G/F architecture and hyperparameters follow the V-JEPA settings in Tables~\ref{tab:network-configuration} and~\ref{tab:training-configuration}. This includes network widths and depths, loss weights, numerical controls, batch size, training seed, and learning rate schedule. The RGB trajectories, action labels, building split, and origin--goal sampling are unchanged. All variants use the epoch 12 checkpoint.

\begin{table}[!t]
\centering
\caption{Visual interfaces of the InJepa encoder variants. All retain native spatial features.}
\label{tab:encoder-interfaces}
\begingroup
\jwjtablebody
\begin{tabular}{@{}llll@{}}
\toprule
Encoder & Input size & Grid & Feature selection \\
\midrule
V-JEPA 2.1 ViT-B & $384\times384$ & $576\times768$ & Native single-frame tokens \\
DINOv3 ViT-B/16 & $256\times256$ & $256\times768$ & Normalized patch tokens \\
CroCo-v2 ViT-Base & $224\times224$ & $196\times768$ & Unmasked image encoder tokens \\
\bottomrule
\end{tabular}
\endgroup
\end{table}

DINOv3 resizes RGB images to $256\times256$ before conversion to floating point and returns normalized patch tokens, excluding class and storage tokens from the Q/A/G/F inputs. CroCo converts RGB to floating point before resizing to $224\times224$ and uses its unmasked single image encoder without cross-view fusion or reconstruction. Both use bilinear resizing with antialiasing, ImageNet normalization, and no crop. V-JEPA retains its $384\times384$ center crop and single frame \texttt{ema\_encoder} interface. Caching and online encoding use identical transforms, detailed in Appendix~\ref{sec:data-evaluation-details}. All encoders remain frozen in evaluation mode.

Grid adaptation changes the fixed spatial position encodings and attention sequence lengths to match each encoder's native tokens. Feature width is unchanged, so no learned feature adapter is added. No spatial pooling or additional whitening is applied, and the Q/A/G/F parameter count remains unchanged. A shared training seed does not guarantee identical initial tensors across variants.

DINOv3 uses four RTX 4090 GPUs, as does V-JEPA. CroCo uses four RTX 3090 GPUs. Both variants apply activation checkpointing throughout the Q/G/F Transformer blocks, whereas V-JEPA retains selected block activations as specified in Appendix~\ref{sec:training-configuration}. This changes the memory and computation tradeoff without changing the architecture or objective. The configuration field \texttt{latent\_dim} denotes feature width $D$, and \texttt{trajectory\_latent\_dim} denotes stochastic dimension $L$. The constructor's \texttt{modes} field specifies a default. Evaluation sets the actual candidate count.

\Needspace{16\baselineskip}
\section{Baseline Reproduction, Deployment, and Compute}
\label{sec:baseline-deployment}

\subsection{Baseline Reproduction}
\label{sec:baseline-reproduction}

We follow training settings disclosed in the original papers. For settings not reported there, we use the official code. Only settings absent from both sources are specified by us. Our data and deployment adaptations, and any paper--code discrepancies, are stated explicitly below. All batch sizes below are global.

\paragraph{Data and implementation scope.}
The shared StreamVLN source does not imply identical filtering, sample multiplicity, or data partitions. InJepa uses 8,004 canonical training trajectories and 2,265 validation trajectories. RAE-NWM retains 15,030 training and 1,208 validation annotation records after length filtering and its own scene split. LWM deduplicates physical trajectories and uses separate subsets for world model/imitation and reinforcement training. NoMaD uses its released weights without StreamVLN training. The 50/11 building split in our experimental setup describes InJepa. Baseline data handling is specified below. Official repository revisions identify the upstream modules, while the disclosed adapters define their use on our data and control interface.

\paragraph{NoMaD.}
We use the official pretrained NoMaD checkpoint~\citep{sridhar2024nomad}, without training or fine tuning on StreamVLN. We bypass the topological image memory, distance based node localization, intermediate node selection and advancement, and the map node arrival test. The final goal image is supplied directly to the policy. The pretrained vision encoder, diffusion model, action decoding, and PD controller are retained. The distance head remains in the checkpoint but is not used for map localization.

Inputs comprise the current RGB image, three previous observations, and the goal image. The history capacity is taken from the released checkpoint's positional encoding. At initialization, the first observation fills the history. Images are resized to $96\times96$ without cropping and normalized with ImageNet statistics. The released model uses EfficientNet-B0, feature width 256, four Transformer layers with four attention heads, and a diffusion U-Net with widths $(64,128,256)$. We use 10 DDPM steps with the squared cosine schedule, eight samples of eight planar waypoints, and explicit goal conditioning. The first sample and its third waypoint are passed to the controller, following the official deployment procedure. Action deltas are rescaled from the normalized range using bounds $(-2.5,-4)$ and $(5,4)$, then cumulatively summed into waypoints. Control limits are specified below.

\paragraph{RAE-NWM.}
We retain the official architecture and objective~\citep{zhang2026rae}. The visual frontend uses DINOv2 with registers at $224\times224$, producing a $16\times16$ grid of 768-dimensional features. The CDiT backbone has 12 blocks, width 768, and 12 heads. The prediction head has two blocks, width 2048, and 16 heads. Training retains velocity prediction on linear flow paths with uniformly sampled times, the official latent normalization statistics, and the pretrained RAE decoder.

\begin{table}[!t]
\centering
\caption{RAE-NWM training settings. The learning rate schedule and evaluated checkpoint are listed separately.}
\label{tab:rae-training}
\begingroup
\jwjtablebody
\begin{tabular}{@{}ll@{}}
\toprule
Setting & Value \\
\midrule
Optimizer / coefficients / offset & AdamW / $(0.9,0.95)$ / $10^{-8}$ \\
Global batch / weight decay / gradient norm limit & 96 / 0 / 1 \\
Learning rate / schedule & $2\times10^{-4}\to2\times10^{-6}$ / linear, 50 epochs \\
EMA decay / training seed & 0.9999 / 42 \\
Context frames / sampled targets per origin & 4 / 4 \\
Target offset range / time normalization & $[-64,64]$ / divide by 128 \\
Precision / workers per process & BF16 mixed / 16 \\
Training hardware / local microbatch / accumulation & 3 RTX 3090 / 8 / 4 \\
Validation / checkpoint interval & 1,000 / 5,000 updates \\
Evaluated checkpoint & EMA at update 155,000 \\
\bottomrule
\end{tabular}
\endgroup
\end{table}

We select the EMA checkpoint at update 155,000 based on the official DreamSim evaluation on the dev set. It achieves the lowest recorded error among the evaluated checkpoints. The 50 epochs specify the learning rate schedule, not the evaluated checkpoint's training duration. Validation uses three batches per process. The global batch of 96 in Table~\ref{tab:rae-training} is preserved through gradient accumulation. Compilation follows the official training command.

The data adapter retains the published StreamVLN RGB frames and their action order. It excludes records shorter than 68 frames: 16,238 of 30,809 records remain, containing 1,868,655 frames. These counts retain annotation multiplicity rather than counting canonical physical trajectories. It integrates recorded commands into planar trajectory metadata using 0.25\,m forward motion and $15^\circ$ turns, starting from a zero pose. These are command derived coordinates, not measured simulator trajectories. Buildings are assigned by a deterministic scene hash: the first SHA-256 byte modulo 10 assigns bucket 0 to validation and the remaining buckets to training. This produces 15,030 training and 1,208 validation records. The official loader forms four-frame contexts and samples four target offsets within $[-64,64]$, excluding the current frame when alternatives exist. Relative planar coordinates are divided by 0.25\,m and normalized using the official distance bounds. Relative times are divided by 128. Images use the official aspect-ratio crop and $224\times224$ resize before RAE preprocessing. Language instructions are not model inputs. The DINOv2 checkpoint, RAE decoder, and latent statistics are required in addition to the trained dynamics checkpoint.

\paragraph{LWM-CroCo.}
We use the official CroCo-v2 frontend, goal fusion, compatibility predictor, policy decoder, and action tokenizer~\citep{wang2026predicting}. Current and goal images are resized to $224\times224$ and normalized with ImageNet statistics. Each image yields $14\times14$ tokens of width 768. The original fusion module is followed by a $768\to384$ projection. Action coordinates use an MLP of widths $3\to512\to384$. Both released decoders use six Transformer layers, width 384, and FFN width 2048. The compatibility predictor has six attention heads and the policy decoder has four. For the policy head count, we retain the released implementation's four heads rather than the six described in the paper. The two codebooks contain 64 action and 64 trajectory centers, respectively. Each trajectory contains at most 63 future waypoints.

\begin{table}[!t]
\centering
\caption{LWM training stages. Both encoder versions use the same optimization settings.}
\label{tab:lwm-training}
\begingroup
\jwjtablebody
\begin{tabular}{@{}lrrrr@{}}
\toprule
Stage & Epochs & Batch & Learning rate & Warmup epochs \\
\midrule
World model & 50 & 8 & $5\times10^{-5}$ & 5 \\
Policy imitation & 40 & 24 & $10^{-4}$ & 4 \\
Policy reinforcement & 40 & 24 & $10^{-4}$ & 4 \\
\bottomrule
\end{tabular}
\endgroup
\end{table}

The official release provides model and inference code. We implement the training stages from the paper. The world model and policy schedules in Table~\ref{tab:lwm-training} follow the reported settings. The paper does not give a separate reinforcement training duration, so our reproduction applies its policy schedule to both policy stages. All stages use Adam with coefficients $(0.9,0.999)$, offset $10^{-8}$, zero weight decay, and seed 0. Learning rates are constant after linear warmup. The remaining adapter settings include four world model candidates per example, log-distance offset 0.01\,m, reinforcement group size 5, sampling temperature 1, KL coefficient 0.01, and clipping coefficient 0.2. The completed imitation policy is the fixed reinforcement reference. Gradients are not propagated to the visual encoder during LWM training.

The Adam coefficients, post-warmup schedule, world model candidate count, distance offset, and reinforcement regularization settings are explicit reproduction choices where neither the paper nor released training code supplies a value. All stages use FP32, with no gradient clipping and four data workers per process. The released decoder dropout is 0.1 and policy position dropout is 0.05. World model squared error is averaged over valid candidate timesteps, and imitation cross entropy is averaged over valid target tokens.

For data conversion, we match each published StreamVLN trajectory to its source episode and replay its recorded actions from the original reset pose to recover aligned positions and orientations. Replay uses 0.25\,m forward motion, $15^\circ$ turns, and sliding. No new RGB frames are rendered for training. Keyframes are retained after at least 0.2\,m of planar displacement. Each training window contains the current image, a sampled future goal image, and up to 63 future keyframes expressed in the current frame's planar coordinates. Short trajectories are padded with a validity mask. Repeated language annotations of the same physical trajectory are deduplicated. Training trajectories are divided by a seeded hash into disjoint world model/imitation and reinforcement halves, containing 3,994 and 3,995 trajectories. Another 2,265 trajectories form validation.

Only trajectories with at least two retained keyframes are eligible. Coordinates are cumulative forward and left displacement in metres and relative yaw in radians, wrapped to $[-\pi,\pi)$. Habitat uses a vertical $Y$ axis and local $-Z$ forward. For seed 0, unique training pose keys are ordered by the SHA-256 hash of \texttt{0:pose\_key}. The first half enters world model/imitation training. Within a window, the goal is sampled uniformly from its valid future keyframes. Each counterfactual draws a different physical trajectory and then a valid origin uniformly. Its motion is expressed relative to that donor origin.

Within each trajectory, valid origins are grouped into bins of at most 63. One origin per bin is used each epoch, cycling without replacement. Imitation uses the fixed epoch 0 pairs. Both codebooks are fitted only on the world model/imitation partition, using k-means++ with 64 centers, one initialization, at most 300 iterations, tolerance $10^{-4}$, and seed 0. World model training combines one factual sequence with three sequences from other trajectories. Its masked squared error targets the log of planar distance to the goal plus 0.01\,m. All three action coordinates are multiplied by 0.1 at the official model input boundary. Distance targets remain in metres. Pseudo labels select the best predicted prefix among the 64 trajectory centers. The policy then uses masked token cross entropy followed by group relative reinforcement learning with world model rewards.

Reinforcement rewards are the negative compatibility values at the final valid waypoints. Advantages are centered and divided by the population standard deviation within each group. Constant-reward groups receive zero advantages. The implementation caps the sequence likelihood ratio against the fixed imitation reference at $1+0.2$. Categorical KL is computed at valid tokens and summed over each sequence, then weighted by 0.01. The resulting objective is averaged over examples and sampled trajectories.

All three stages complete their scheduled epochs. LWM-CroCo uses the final world model and reinforcement policy, after 32,700 and 8,680 updates, respectively. Imitation completes 8,720 updates. World model and imitation training use two RTX 3090 GPUs. Reinforcement training uses one RTX 3090 with microbatch 8 and three accumulation steps, retaining global batch 24.

\paragraph{LWM-VJEPA.}
Only the single-image visual frontend is replaced by V-JEPA~2.1. Its official preprocessing resizes the short side to 438 pixels and center crops to $384\times384$, yielding $24\times24$ tokens of the same width 768. We update the spatial position indices and policy visual token count from 196 to 576. All other model, data, training, and planning settings are identical to LWM-CroCo. This version is trained on two RTX 4090 GPUs with the same global batches and stage durations. The V-JEPA input contains one frame per observation. The CroCo fusion module, feature projection, action tokenizer, and fitted codebooks are retained. The world model and both policy stages are trained for this frontend. Weights are not obtained by swapping encoders only at evaluation.

\subsection{Control Interfaces and Planning Budgets}

\begin{table}[!t]
\centering
\caption{Control interfaces. Motion limits apply to one environment control step.}
\label{tab:control-interfaces}
\begingroup
\jwjtablebody
\begin{tabular}{@{}llll@{}}
\toprule
Method & Translation limit & Rotation limit & Execution unit \\
\midrule
InJepa & $0.25$\,m & $15^\circ$ & One discrete action \\
RAE-NWM & $0.25$\,m & $15^\circ$ & $1$\,s integration \\
LWM-CroCo / LWM-VJEPA & $0.25$\,m & $15^\circ$ & $1$\,s integration \\
NoMaD & $0.20$\,m & $0.4$\,rad ($22.92^\circ$) & $1$\,s command hold \\
\bottomrule
\end{tabular}
\endgroup
\end{table}

Following the closed loop planning and execution procedures in prior work~\citep{shen2026efficient,zhang2026rae,sun2026intact}, our evaluation uses a common replanning protocol: each method executes one environment control step per plan and replans from the next observation. All methods share the Clean150 tasks, sensor configuration, 200-step interaction budget, and 1\,m arrival radius. Table~\ref{tab:control-interfaces} specifies their control interfaces.

InJepa executes either a forward movement or an in-place turn. RAE-NWM uses continuous control, and LWM converts its first selected waypoint to continuous control. For both methods, planar displacement is clipped to a radius of 0.25\,m and yaw to $\pm15^\circ$. Translation and rotation can occur together. Thus equal step limits do not imply identical motion capabilities. With no collisions and every step used for forward motion, 200 steps permit at most 50\,m for InJepa, RAE-NWM, and LWM, or 40\,m for NoMaD. Actual travel depends on turns, commands, and collisions. The continuous controllers allow at most 200\,s of nominal simulated execution. This excludes planning time, and the discrete InJepa interface has no assigned physical duration per step.

NoMaD retains the official PD controller's internal 0.25\,s timestep and waypoint scale of $0.2/4=0.05$. Its linear and angular velocities are bounded by 0.2\,m/s and 0.4\,rad/s, respectively. Each command is held for 1\,s before the next observation.

\begin{table}[!t]
\centering
\caption{Planning budgets. Counts and horizons refer to one planning call.}
\label{tab:planning-budgets}
\begingroup
\jwjtablebody
\begin{tabular}{@{}llp{7.4cm}@{}}
\toprule
Method & Candidates / horizon & Generation and selection \\
\midrule
InJepa-VJEPA & $8$ / $4$ & Recurrent Q/G/F rollout with complete branch scoring. \\
RAE-NWM & $120$ / $8$ & One CEM update with top 3 candidates and 10 Euler points per prediction. \\
LWM & $32$ / up to $63$ & Policy proposals with compatibility scoring of each valid trajectory. \\
NoMaD & $8$ / $8$ & 10 diffusion steps, first sample, waypoint index 2. \\
\bottomrule
\end{tabular}
\endgroup
\end{table}

Table~\ref{tab:planning-budgets} summarizes the planning budgets. RAE-NWM retains its curve parameterization of action sequences and terminal DINO feature scoring. LWM uses a 64-token capacity including the initial position. A termination token determines valid trajectory length. NoMaD follows the official choice of the first sampled trajectory and its third waypoint, without an additional candidate scorer. These budgets preserve the respective deployment procedures and do not constitute an equal compute comparison.

\subsection{Euler Discretization}
\label{sec:euler-discretization}

Reducing RAE-NWM from 50 to 10 Euler points changes the numerical solver for future latent generation. It retains the trained checkpoint, action conditioned dynamics, CEM search, eight-step rollout, and goal feature scoring. The two grids have 49 and 9 integration intervals, respectively. This adjustment preserves the planning mechanism but can change predicted features and candidate rankings.

An isolated RTX 4090 profile of the original 50-point setting recorded 206.45\,s for the first planning call and 82.02\,s on average for the following three calls. Candidate rollouts and the final mean-trajectory rollout took 77.61\,s and 4.20\,s, respectively, accounting for 99.75\% of the latter average. The original setting therefore incurred substantial model inference cost, motivating the reduced integration budget.

Euler50 completed seven tasks in a partial evaluation on Clean150. One task reached the goal, giving Reach SR of 14.29\% and ReachSPL of 0.074790. Two interrupted tasks are excluded. The 50-point setting also shows limited navigation performance on this subset.

Euler10 seeds 1 and 2 also skip the final mean-trajectory visualization rollout and RGB decoding when these outputs are unused for action selection. Checks with the same checkpoint and inputs confirmed identical selected actions and random generator states before and after this execution optimization. These checks concern implementation equivalence, not navigation equivalence between the two Euler grids.

\subsection{Observed Planning Latency}

\begin{table}[!t]
\centering
\caption{Planning latency on one dedicated RTX 3090. Statistics exclude the first five calls. Times are in milliseconds.}
\label{tab:planning-latency}
\begingroup
\jwjtablebody
\begin{tabular}{@{}lrrrr@{}}
\toprule
Method & Measured calls & Mean & Median & P95 \\
\midrule
NoMaD & 30 & 106.13 & 106.17 & 107.54 \\
LWM-CroCo & 20 & 413.12 & 358.91 & 655.63 \\
LWM-VJEPA & 30 & 611.55 & 494.03 & 921.40 \\
RAE-NWM (Euler10) & 10 & 28,357.95 & 28,360.16 & 28,383.00 \\
\midrule
InJepa-CroCo & 30 & 527.60 & 527.31 & 545.02 \\
InJepa-DINOv3 & 30 & 568.38 & 572.68 & 584.16 \\
InJepa-VJEPA & 18 & 1,421.23 & 1,422.28 & 1,456.28 \\
\bottomrule
\end{tabular}
\endgroup
\end{table}

Table~\ref{tab:planning-latency} reports a short timing run on the same Clean150 task (index 0), with one method running at a time on a dedicated RTX 3090. GPU monitoring detected no other compute processes. CUDA is synchronized around each policy call. Timing includes visual encoding, planning, and action return, but excludes simulator stepping. We discard the first five calls as warmup and report the mean, median, and 95th percentile of the remaining calls. Runs allow up to 35 calls, or 15 for RAE-NWM, and end upon arrival. Measured counts therefore differ. NoMaD uses its map-free deployment, and InJepa uses full feedback.

These measurements describe the evaluated deployments on a single task. Subsequent observations differ across methods, so they do not estimate average latency across scenes. The separate RTX 4090 Euler50 profile in Appendix~\ref{sec:euler-discretization} uses a different protocol and is not included in this table. Table~\ref{tab:navigation-results} compares complete navigation systems. The controlled mechanism comparison instead holds model weights, control interfaces, and candidate budgets fixed within InJepa.

\section{Data and Evaluation Details}
\label{sec:data-evaluation-details}

\subsection{Training Data and Visual Features}

We use the published StreamVLN R2R and RxR RGB and action trajectories~\citep{wei2025streamvln}. The building split contains 8,004 training trajectories from 50 scenes and 2,265 validation trajectories from 11 disjoint scenes. These trajectories contain 648,317 and 201,238 cached RGB observations, respectively. Language instructions are not model inputs.

The data contain 10,269 canonical trajectories. Annotation row and origin/goal indices are zero based. Temporal indices refer to numeric frame order, rather than JPEG filename numbers. Baseline filtering and partition rules are specified in Appendix~\ref{sec:baseline-reproduction}.

\paragraph{Visual preprocessing.}
All three encoders remain frozen and produce spatial features of width 768. V-JEPA 2.1 uses its official deterministic preprocessing: bilinear resizing of the shorter side to 438 pixels with antialiasing, followed by a $384\times384$ center crop. Each RGB observation is encoded separately, with a temporal input length of one. DINOv3 uses bilinear resizing to $256\times256$ on uint8 images before conversion to floating point. CroCo-v2 converts RGB to floating point before bilinear resizing to $224\times224$. Both use antialiasing and no crop. All inputs use channel means $(0.485,0.456,0.406)$ and standard deviations $(0.229,0.224,0.225)$ after scaling to $[0,1]$.

The frozen weights are V-JEPA 2.1 ViT-B distilled from ViT-G at resolution 384, DINOv3 ViT-B/16 pretrained on LVD-1689M, and CroCo-v2 ViT-Base. We retain $576$, $256$, and $196$ spatial tokens, respectively. DINOv3 uses normalized patch tokens from the encoder output, while CroCo uses its single image encoder without masking. Features are cached in float32, with no spatial pooling or additional whitening. Training and online evaluation use the same preprocessing for each encoder.

\paragraph{Training samples.}
A sample uses up to four factual observations and their aligned actions. For each nonterminal origin, Q uses the trajectory endpoint as a goal and, when available, one deterministically hashed future goal at least four steps ahead. Duplicate goals are removed. Thus the goal sampling horizon does not restrict goals to the next four steps. Q is supervised only on the next observed grid. The training set contains 640,313 nonterminal origins and 1,218,429 distinct origin--goal occurrences. Beginning and terminal actions, history masks, and gradient boundaries follow Appendix~\ref{sec:implementation-details}.

\paragraph{Offline subset.}
The fixed diagnostic subset contains 4,096 origin--goal pairs from 265 validation trajectories. Validation pairs use the same endpoint and hashed future goal construction described above. Selection balances scenes and data sources, caps each trajectory at 16 pairs, and uses deterministic hashing. The same subset is used across training checkpoints and loss ablations. Histories, reference future grids, and action labels come from the recorded trajectories. Q alignment excludes steps beyond the sampled goal. F prediction uses recorded actions and excludes steps beyond the trajectory endpoint. F and G Local count each distinct origin once, while goal conditioned diagnostics retain distinct origin--goal pairs. Prediction errors are aggregated over valid samples at each horizon. Four step summaries give equal weight to the four horizon means. Evaluation seeds change prior samples, not the validation split or trained model.

\begin{table}[!t]
\centering
\caption{Clean150 scene IDs and task counts.}
\label{tab:clean150-scenes}
\begingroup
\jwjtablebody
\begin{tabular}{@{}lr@{\hspace{3em}}lr@{}}
\toprule
MP3D scene ID & Tasks & MP3D scene ID & Tasks \\
\midrule
\texttt{2azQ1b91cZZ} & 15 & \texttt{Z6MFQCViBuw} & 15 \\
\texttt{8194nk5LbLH} & 15 & \texttt{oLBMNvg9in8} & 15 \\
\texttt{EU6Fwq7SyZv} & 14 & \texttt{pLe4wQe7qrG} & 7 \\
\texttt{QUCTc6BB5sX} & 12 & \texttt{x8F5xyUWy9e} & 13 \\
\texttt{TbHJrupSAjP} & 15 & \texttt{zsNo4HB9uLZ} & 15 \\
\texttt{X7HyMhZNoso} & 14 & Total & 150 \\
\bottomrule
\end{tabular}
\endgroup
\end{table}

\subsection{Clean150 Task Poses}
\label{sec:clean150-poses}

Clean150 uses Habitat's MP3D scene assets, stored as \texttt{mp3d/<scene>/<scene>.glb}. Its 11 buildings match the MP3D validation scene identities in the official Habitat PointNav data and are disjoint from our training and validation buildings. The 150 image goal tasks are fixed project tasks, not the official PointNav episode list. Table~\ref{tab:clean150-scenes} lists their scene IDs and counts.

\paragraph{Task construction and screening.}
We generate 200 initial tasks with Habitat's official PointNav generator and random seed 20260910. Start and goal positions are connected on the navigation mesh, with geodesic distance in $[5,10)$\,m, geodesic to Euclidean distance ratio at most 1.3, and cumulative horizontal path turning at most $120^\circ$. Both endpoints have at least 0.2\,m clearance from the navigation mesh boundary. Start and goal orientations follow the initial and final approach directions of the shortest path, respectively, using reference points 1\,m along the path from each endpoint. An initial forward probe checks local traversability, and a reference follower must reach the goal within 200 steps with cumulative turning at most $180^\circ$. Start and goal images pass a black pixel check before manual inspection of the images and reference routes. This inspection excludes 33 tasks with severe reconstruction defects or uninformative close views.

From the remaining 167 tasks, we select 150 under balanced building quotas and fixed distance quotas. Five consecutive 1\,m distance bins contain 8, 22, 38, 52, and 30 tasks, respectively. Selection minimizes deterministic costs derived from episode ID hashes subject to these quotas. Selected tasks retain their original positions, orientations, and order.

These checks focus evaluation on the goal directed navigation mechanisms studied here and improve control over comparisons across methods. The initial distance range limits disproportionate effects from very short or long tasks. Constraints on detour ratio and path turning reduce the influence of complex topological exploration, focusing evaluation on sustained image goal navigation and multistep closed loop planning. The goal point must also lie within the initial camera frustum with an unobstructed line of sight. This reduces the influence of global goal search and focuses evaluation on sustained planning and action realization given a visual goal. The reference follower only checks reachability within the prescribed interaction budget and does not guide the evaluated policies. Image screening targets only severe reconstruction artifacts and uninformative close views to reduce the influence of reconstruction quality on navigation metrics. Low texture or minor mesh defects alone do not trigger exclusion. Neither task construction nor final selection uses evaluation results from any learned policy. All methods share the fixed tasks, sensor settings, 200-step interaction budget, and 1\,m arrival radius, providing a common evaluation basis for comparisons within the specified task distribution.

\paragraph{Positions and orientations.}
Each task is specified by its task index, episode ID, scene ID, start position, start orientation, goal position, goal image orientation, success radius, and initial geodesic distance. Positions are $(x,y,z)$ in metres in the native Habitat scene coordinates. Orientations are unit quaternions in $(x,y,z,w)$ order. Each reset uses the recorded start pose. The goal image is rendered at the recorded goal position and orientation. The goal orientation defines the image, not an orientation constraint for arrival. Recorded poses preserve the original numerical precision and are identical across evaluation seeds.

For example, task 0, \texttt{2azQ1b91cZZ\_000}, starts at $(18.343557,0.127110,7.543764)$ with quaternion $(0,0.869414,0,0.494085)$. Its goal is $(12.536557,0.127110,10.668974)$ with image quaternion $(0,0.848889,0,0.528572)$. These displayed values are rounded.

\subsection{Simulator and Control Settings}

We use Habitat-Lab and Habitat-Sim 0.2.4. RGB observations have resolution $640\times480$, horizontal field of view $79^\circ$, and camera offset $(0,1.25,0)$\,m relative to the agent. Agent height and radius are 1.5\,m and 0.1\,m. Sliding is enabled and physics simulation is disabled. InJepa uses forward motion of 0.25\,m or left/right turns of $15^\circ$. Each planning round executes one environment control step before acquiring a new observation and replanning. The task budget is 200 such steps, including turns. A task succeeds when geodesic distance falls below 1\,m. Mean initial geodesic distance is 7.932959\,m.

The policy receives RGB history, previous actions, and the goal image. Simulator pose and geodesic distance are used for evaluation, not as policy inputs. The default primary model uses $K=8$, $H=4$, and $\lambda_{\mathrm c}=1$. We select these planning hyperparameters using four dev metric groups: Q Future Alignment, F Prediction Error, G Local/G Goal accuracy, and Intent Realization Error. The default configuration is fixed before inspecting Clean150 results. Our hyperparameter search is limited in scope. The experiments prioritize evaluating the intention first architecture over exhaustive performance optimization. Clean150 is used to report navigation performance in unseen scenes. Dev metrics provide complementary analysis of internal model behavior. Navigation repeats use evaluation seeds 0, 1, and 2 on the same 150 tasks. They measure evaluation sampling variability, not independent training repeats or additional test tasks. Complete branch scoring and deployment definitions are specified in Appendices~\ref{sec:scoring-details} and~\ref{sec:deployment-mechanisms}. Metric formulas and aggregation rules are given in Appendix~\ref{sec:evaluation-metrics}.

\end{document}